\documentclass{article}

\PassOptionsToPackage{numbers, compress}{natbib}

\newif\ifincludesection
\includesectiontrue

\usepackage[preprint]{neurips_2025}

\usepackage[utf8]{inputenc} 
\usepackage[T1]{fontenc}    
\usepackage{url}            
\usepackage{booktabs}       
\usepackage{amsfonts}       
\usepackage{nicefrac}       
\usepackage{microtype}      
\usepackage[dvipsnames]{xcolor}
\usepackage{enumitem}
\usepackage{minitoc}
\usepackage{graphicx}
\usepackage{svg}

\usepackage{mathtools}
\usepackage{amsmath,amssymb, amsthm}

\usepackage{caption} 
\DeclareMathOperator{\tr}{Tr}

\newtheorem{assumption}{Assumption}
\newtheorem{lemma}{Lemma}

\usepackage{algorithm}
\usepackage{algpseudocode}

\theoremstyle{plain}

\newtheorem{remark}{Remark}

\newtheorem{definition}{Definition}

\newcommand{\R}{\mathbb R}
\newcommand{\SO}{\mathbb S}
\newcommand{\M}{\mathcal M}
\newcommand{\C}{\mathcal C}
\newcommand{\UB}{\mathcal O}
\newcommand{\VC}{\mathbb V}
\newcommand{\F}{\mathcal F}
\newcommand{\xs}{x_\star}
\newcommand{\fs}{f_\star}
\newcommand{\Df}{\nabla f}

\newcommand{\VCf}{\VC_\ell}

\newcommand{\rhostar}[1]{\rho^\star(#1)}
\newcommand{\rhostarK}[1]{\rho_K^\star(#1)}
\newcommand{\ml}{(\mu, L)}
\newcommand{\CGD}{\mathrm{CGD}}
\newcommand{\EF}{\mathrm{EF}}
\newcommand{\EFtw}{\mathrm{EF^{21}}}

\newcommand{\Pp}{(P,p)}
\newcommand{\Vbase}{\mathcal V}
\newcommand{\Lyappp}[3]{\Vbase_{\Pp}(#1,#2;#3)}
\newcommand{\Lyap}[3]{\Vbase(#1,#2;#3)}

\newcommand{\Fml}{\F_{\mu, L}}
\newcommand{\Mfxxi}[3]{\M(#1,#2;#3)}
\newcommand{\MfxxiK}[3]{\M^K(#1,#2;#3)}

\usepackage[
  colorlinks=true,
  linkcolor=blue,      
  citecolor=green!50!black,  
  filecolor=magenta,
  urlcolor=cyan
]{hyperref}    

\usepackage{cleveref}

\crefname{figure}{figure}{figures}
\Crefname{figure}{Figure}{Figures}
\crefname{table}{table}{tables}
\Crefname{table}{Table}{Tables}

\crefname{section}{section}{sections}
\Crefname{section}{Section}{Sections}
\crefname{subsection}{subsection}{subsections}
\Crefname{subsection}{Subsection}{Subsections}

\crefname{equation}{equation}{equations}
\Crefname{equation}{Eq.}{Eqs.}

\crefname{theorem}{theorem}{theorems}
\Crefname{theorem}{Theorem}{Theorems}
\crefname{lemma}{lemma}{lemmas}
\Crefname{lemma}{Lemma}{Lemmas}
\crefname{proposition}{proposition}{propositions}
\Crefname{proposition}{Proposition}{Propositions}
\crefname{corollary}{corollary}{corollaries}
\Crefname{corollary}{Corollary}{Corollaries}
\crefname{definition}{definition}{definitions}
\Crefname{definition}{Definition}{Definitions}
\crefname{remark}{remark}{remarks}
\Crefname{remark}{Remark}{Remarks}
\crefname{example}{example}{examples}
\Crefname{example}{Example}{Examples}
\Crefname{assumption}{Assumption}{Assumptions}

\crefname{algorithm}{algorithm}{algorithms}
\Crefname{algorithm}{Algorithm}{Algorithms}

\usepackage{thmtools}
\usepackage{thm-restate}

\newenvironment{proofsketch}{%
  \proof}{\endproof}
\input{certified.tex}

\usepackage{graphbox}

\title{Tight analyses of first-order methods with error feedback}

\newcommand{\inria}{INRIA, D.I. École Normale Supérieure, PSL Research University, 75005 Paris, France}
\newcommand{\cmap}{CMAP, CNRS, École polytechnique, Institut Polytechnique de Paris, 91120 Palaiseau, France}

\author{%
    \textbf{Daniel Berg Thomsen}\(^{1, 2}\)\thanks{Correspondence to \texttt{daniel.berg-thomsen@inria.fr}} \qquad Adrien Taylor\(^{1}\) \qquad Aymeric Dieuleveut\(^{2}\) \\
    \(^{1}\)\inria \\
    \(^{2}\)\cmap
}

\renewcommand{\leq}{\leqslant}
\renewcommand{\geq}{\geqslant}
\renewcommand{\succeq}{\succcurlyeq}
\renewcommand{\preceq}{\preccurlyeq}
\tcbuselibrary{breakable} 

\renewcommand{\makecertificatenotice}{%
  \vspace*{\certnoticevspace}%
  \begin{center}
    \emph{Proof certificates are annotated using\footnote{We generate certificates of validity using a Computer Algebra System (CAS, via WolframScript) and Performance Estimation Problems (PEP). CAS annotations mark partial verification of proof identities, whereas PEP annotations indicate full verification of the statement. Each annotation links to the corresponding file in our GitHub repository}}%
    \par
    \includegraphics[height=\certnoticeheight]{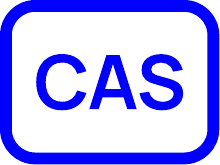}\hspace{\certnoticesep}%
    \includegraphics[height=\certnoticeheight]{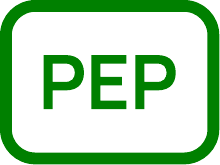}%
  \end{center}%
}

\begin{document}

\maketitle


\begin{abstract}
    Communication between agents often constitutes a major computational bottleneck in distributed learning.
    One of the most common mitigation strategies is to compress the information exchanged, 
    thereby reducing communication overhead. To counteract the degradation in convergence
    associated with compressed communication, error feedback schemes---most notably $\EF$
    and $\EFtw$---were introduced. In this work, we provide a \emph{tight analysis} of both 
    of these methods. Specifically, we find the 
    Lyapunov function that yields the best possible convergence rate for each method---with
    matching lower bounds.  This principled approach yields sharp performance guarantees 
    and enables a rigorous, 
    apples-to-apples comparison between $\EF$, $\EFtw$, and compressed gradient descent. Our 
    analysis is carried out in the simplified single-agent setting, which allows for 
    clean theoretical insights and fair comparison of the underlying mechanisms.
\end{abstract}

\begin{caspara}
To consolidate and support  our theoretical results, we complement each theoretical statement with analytical or numerical validation. Specifically, we provide certificates of correctness generated either with a \textit{Computer Algebra System} (CAS), using a WolframScript, for symbolic verification, or using \textit{Performance Estimation Problems (PEP)} for numerical validation. CAS enable verification of algebraic identities, while PEP annotations indicate numerical validation of  complete statements. These certificates are highlighted in the paper using
\includegraphics[height=1em]{CAS.pdf} and \includegraphics[height=1em]{PEP.pdf} markers, which are direct links to the corresponding Jupyter notebook or WolframScript in our public GitHub repository.
\footnote{While these certificates do not replace the mathematical proofs presented in the paper, they serve as an additional layer of transparency and error checking, analogous to unit tests in software development. This practice provides a reproducible, independently verifiable basis for our theoretical claims, thereby reducing the risk of oversights in complex derivations.}
\end{caspara}

\section{Introduction}
Over the past decade, distributed optimization has become a cornerstone of large-scale machine 
learning. This shift is driven by major increases in the size of models and training data,
as well as increasing societal concerns about data ownership and privacy. 
Ultimately, solutions in which training is distributed across a network of $n$ 
agents, each retaining its own local data, under the coordination of a central server, 
have emerged as one of the most natural and efficient solutions to this problem~\cite{mcmahan_communication-efficient_2017,kairouz_advances_2019}. 
Formally, the goal is to solve the following minimization problem:
\begin{equation}
 \min _{x\in \mathbb R^d} \left\{f(x ):=\frac{1}{n} \sum_{i=1}^n f_i(x)\right\}.
\end{equation}
Classical methods such as distributed gradient descent and its stochastic variants achieve 
linear speedups in iteration complexity with respect to the number of agents. However, they 
often suffer from significant \emph{communication overhead}, as gradients or model updates must be exchanged 
frequently over bandwidth-limited channels~\citep{seide_1-bit_2014,chilimbi_project_2014,strom_scalable_2015}. 
As the scale of models keep increasing, this communication bottleneck has been identified 
early on as a critical limitation, prompting the development of methods aimed at reducing communication costs. 
Two main strategies are favored: scarcely communicating with the central server, known as 
\textit{local iterations}~\citep[see e.g.~][]{mcmahan_communication-efficient_2017,karimireddy_scaffold_2020}---and
transmitting \emph{compressed updates}, which aim to reduce the size of the exchanged information. 
Compression mechanisms can be applied to reduce communication either from agents to the 
server~\citep{seide_1-bit_2014,alistarh_qsgd_2017,alistarh_convergence_2018,mishchenko_distributed_2019,karimireddy_error_2019,wu_error_2018,horvath_natural_2022,li_acceleration_2020,richtarik_ef21_2021} 
or from the server to the agents~\citep{harrane_reducing_2018,tang_doublesqueeze_2019,liu_double_2020,zheng_communication-efficient_2019,philippenko_artemis_2020,philippenko_preserved_2021,gorbunov_linearly_2020,sattler_robust_2019,fatkhullin_ef21_2021}. 
This paper focuses on methods using compression operators, which encompass a variety of 
strategies, including selecting only a fraction of the weights to be transmitted (e.g.,~the top 
\(K\) coordinates~\cite{alistarh_convergence_2018}) or communicating low-precision 
updates via quantization~\cite{alistarh_qsgd_2017}. 

\begin{algorithm}[h]
    \caption{Compressed gradient descent (CGD)}
    \label{alg:cgd}
    \begin{algorithmic}[1]
        \State{\bfseries initialization:} \(x_0 \in \R^d, \eta > 0\)
        \For{\(k = 0, 1, 2, \ldots, N\)}
            \State Agent $i\in [n]$ compresses \(\Df_i(x_k)\) and communicates \(m_k^{(i)} \coloneqq \C(\Df_i(x_k))\) 
            \State Server updates \(x_{k+1} \gets x_k - \eta \cdot \frac{1}{n} \sum_{i = 1}^n m_k^{(i)}\)
        \EndFor
    \end{algorithmic}
\end{algorithm}
Formally, a compression operator is a possibly random mapping $\C: \mathcal X \to \mathcal X$, such that $\C(X)$ can be encoded (almost surely or on average) with a lower number of bits than $X$.
The most natural algorithm leveraging communication compression with a centralized server is the \emph{compressed gradient descent} algorithm ($\CGD$), which is described in Algorithm~\ref{alg:cgd}. 
The main idea is to perform a distributed gradient step, with the compression operator \(\C\) applied to the gradient of each agent before communication. 
Although this compression scheme reduces the communication cost, it comes at the expense of 
non-convergence in any practical setting~\cite{beznosikov_biased_2020}. 

To assess the general impact of compression schemes on the rate 
of convergence, one typically leverages the fact that these compressors all satisfy generic assumptions. 
These include unbiasedness, i.e., \(\mathbb E[\C(x)]= x\) for any \(x\in \mathcal X\), together with relatively bounded variance, which states that 
\(\mathbb E[\|\mathcal C(x)-x\|^2]\le \omega \|x\|^2\) for any \(x\in \mathcal X\)~\citep{alistarh_qsgd_2017,wu_error_2018,mishchenko_distributed_2019,chraibi_distributed_2019,gorbunov_unified_2020,reisizadeh_fedpaq_2020,horvath_natural_2022,kovalev_lower_2021,philippenko_artemis_2020,philippenko_preserved_2021,haddadpour_federated_2021,li_canita_2021,khirirat_distributed_2018}, 
or \textit{contractiveness}~\cite{stich_sparsified_2018,ivkin_communication-efficient_2019,koloskova_decentralized_2019,gorbunov_linearly_2020,beznosikov_biased_2020}, 
defined as follows:
\begin{assumption}[Contractive compression operator]\label{as:compression}
    The compression operator \(\C\) is a stochastic operator such that, for some \(\epsilon \in [0, 1)\),
    \begin{equation}
         \text{for  all} \quad x \in \mathbb{R}^d, \quad \mathbb{E} \left[\|x - \C(x)\|^2\right] \leq \epsilon \|x\|^2 . 
    \end{equation}
\end{assumption}

The standard way to improve  $\CGD$ is to leverage the asymmetry of information: each agent has access 
to the exact gradient before compression and can therefore track the discrepancy between 
the exact gradient and the transmitted (compressed) message. This discrepancy can be stored 
and used as a correction term in subsequent iterations—a principle that lies at the heart 
of \textit{error feedback techniques}. The most basic mechanism used is known as 
\emph{classic error feedback} (EF), where each agent stores the difference between the true 
gradient and its compressed version locally, and incorporates this error into the next round of 
communication. This method,
outlined in Algorithm~\ref{alg:ef}, was first introduced in~\cite{seide_1-bit_2014} and 
later analyzed in~\cite{stich_sparsified_2018,wu_error_2018,karimireddy_error_2019,alistarh_convergence_2018,stich_error-feedback_2020}. 
Notably, this method converges in many practical settings, effectively addressing the problem of non-convergence for $\CGD$.

\begin{algorithm}[h]
    \caption{Classic error feedback (EF)}
    \label{alg:ef}
    \begin{algorithmic}[1]
        \State{\bfseries initialization:} \(x_0 \in \R^d, \eta > 0, e^{(i)}_0 = 0\) for \(i = 1, \ldots, n\)
        \For{\(k = 0, 1, 2, \ldots, N\)}
            \State Agent \(i\in [n]\) compresses \(e_k^{(i)} + \eta \nabla f_i(x_k)\) and communicates \(m_k^{(i)} \coloneqq \C(e_k^{(i)} + \eta \nabla f_i(x_k))\)
            \State Agent \(i\in [n]\) updates \(e_k^{(i)} \gets e_k^{(i)} + \eta \nabla f_i(x_k) - \C(e_k^{(i)} + \eta \nabla f_i(x_k))\)
            \State Server updates \(x_{k+1} \gets x_k - \frac{1}{n}\sum_{i=1}^n m_k^{(i)}\)
        \EndFor
    \end{algorithmic}
\end{algorithm}

More recently, a variant of the classic error feedback mechanism, known as 
 $\EFtw$ was introduced by~\citet{richtarik_ef21_2021}, and is presented in 
Algorithm~\ref{alg:ef21}. Unlike classic error feedback, $\EFtw$ focuses on communicating 
a gradient estimate that is more robust to the variance observed in gradients received 
from \emph{different} agents
around the minimum of finite sum objectives. This method has since been extended in several 
directions~\cite[e.g.][]{fatkhullin_ef21_2021,fatkhullin2023momentum,gruntkowska_ef21-p_2022,makarenko2022adaptive}.
\begin{algorithm}[h]
    \caption{Error Feedback 21 --- $\EFtw$}
    \label{alg:ef21}
    \begin{algorithmic}[1]
        \State{\bfseries initialization:} \(x_0 \in \R^d;\) 
        step size \(\eta > 0;\) 
        \(d^{(i)}_0 = \C(\Df_i(x_0))\) for \(i = 1, \ldots, n;\)
        \For{\(k = 0, 1, 2, \ldots, N\)}
            \State Server updates \(x_{k+1} \gets x_k - \eta \cdot \frac{1}{n} \sum_{i=1}^n d^{(i)}_k\)
            \State Agent \(i\in [n]\) compresses \(\Df_i(x_{k+1}) - d^{(i)}_k\) and communicates \(m^{(i)}_k := \C(\Df_i(x_{k+1}) - d^{(i)}_k)\)
            \State Agent \(i\in [n]\) updates \(d^{(i)}_{k+1} \gets d^{(i)}_k + m^{(i)}_k\)
        \EndFor
    \end{algorithmic}
\end{algorithm}

Error feedback techniques are widely regarded as highly effective, and $\EF$ was described as 
\textit{``compression for free''} as early as 2019~\cite{karimireddy_error_2019}. Despite that,  and the
abundant literature on the topic, the
precise impact of error feedback techniques on performance remains difficult to assess. 
Comparison is complicated by the diversity of settings under which methods
are analyzed: different function classes 
(smooth, convex, or nonconvex), a range of algorithmic enhancements (acceleration, adaptivity, 
variance reduction, etc.), and a variety of performance measures (different Lyapunov 
functions)~\cite{fatkhullin_ef21_2021,richtarik2022threepoint,makarenko2022adaptive,gruntkowska_ef21-p_2022,zhao2022beer,wang2022communication,dorfman2023docofl}.
While some works provide insightful counter-examples—e.g., \citet{beznosikov_biased_2020} show 
that classic error feedback effectively addresses the limitations of  $\CGD$ in distributed 
settings—many others simply propose a Lyapunov function and establish an upper bound 
without demonstrating its tightness. As a result, claims about “compression for free” 
are often based on comparisons between potentially loose guarantees, which may not 
reliably reflect real algorithmic performance. 
The length and complexity of the proofs involved typically make it difficult to ensure the tightness of the results, and most proofs are constructed in an ad hoc manner.
Consequently, it is difficult to determine which methods are actually worst-case optimal 
based on upper bounds whose tightness is not always assured.


As a result, even remarkably simple questions remain only partially answered:
\begin{tcolorbox}\centering
What is the optimal convergence rate that each method can attain? \\ 
Given an optimization setting, what method should we choose? \\
How should each method be optimally tuned? 
\end{tcolorbox}
Our goal is to provide definitive answers to parts of these questions.
In this paper, we take a complementary perspective to the existing literature and offer
a tight, principled comparison of the three methods. Specifically, we derive their optimal
tuning, identify an optimal Lyapunov function for each method, and compute the \emph{exact}
optimal convergence rate for \emph{any} Lyapunov function within our class of candidate Lyapunov functions.

To make this comparison sharp and transparent, we adopt a deliberately simple yet 
representative setup: we consider smooth and strongly convex functions in the 
single-agent setting ($n = 1$). While simple, this regime is widely recognized as a 
crucial stepping stone—not only for building intuition, but also as its own theoretical
contribution~\citep[e.g.,][]{stich_sparsified_2018,karimireddy_error_2019}. 
The single-agent setting is also of independent interest: in the field of \emph{sparsity-aware} neural-network training, sparse-update methods have been shown to correspond to error feedback~\cite{lin2020dynamic}.
In this context, \emph{tightness} means that we identify the best possible Lyapunov 
function within a given class \emph{and} compute the exact worst-case convergence rate over 
the class of problems considered.

Our methodology draws on the \emph{performance estimation}  
framework~\cite{drori2014contributions,taylor2017exact}, which enables the numerical 
derivation of exact convergence rates for a wide range of first-order methods. In 
particular, recent advances~\cite{taylor2018lyapunov,upadhyaya2023automated} demonstrate 
how to automatically search for optimal Lyapunov functions. While these approaches are 
primarily numerical, we build upon insights from their underlying proof 
structures~\cite{goujaud2023fundamental} to derive new analytical results. 
  

\paragraph{Contributions.} This work makes the following contributions:
\begin{enumerate}[leftmargin=*, itemsep=0.5pt,topsep=0pt]
\item From a methodological perspective, this paper shows how to apply the performance estimation framework to algorithms from the \emph{federated learning} literature that incorporate compression schemes. By leveraging this methodology—both analytically and numerically—this work paves the way for a more precise and reliable understanding of federated and distributed learning methods.
\item This work provides a \emph{tight} analysis of $\EF$ and $\EFtw$, and compare them with compressed gradient descent in the single-agent setting, on \(L\)-smooth, \(\mu\)-strongly convex functions. In particular, we give an analytical formula of the best possible contraction rate, by analyzing an optimal Lyapunov function within a class of candidate Lyapunov functions defined in Definition~\ref{def:candidate_lyapunov}. Furthermore, we provide the optimal tuning for the step size in both those algorithms.
\item We demonstrate that those rates are achieved, proving that the analysis is tight. 
\item We conclude that the complexities of $\EF$ and $\EFtw$ are perfectly identical in this particular setting. Moreover, $\CGD$ outperforms both methods—both in terms of the range of settings where it converges and in terms of the optimal convergence rate achieved.
\item Finally, we contribute to the process of deriving \emph{simple} Lyapunov functions for first-order methods, and extend known results for fixed-step methods to the setting of methods using compression.
\end{enumerate}

\paragraph{Paper outline.} The rest of the paper is organized as follows.
    In Section~\ref{sec:background}, we provide background on the relevant existing results
    for  $\CGD$, $\EF$, and $\EFtw$. We also provide the necessary background on the techniques from
    the performance estimation literature  needed to outline the methodology we use,
    as well as the definition of the classes of Lyapunov functions used.
    Section~\ref{sec:main_results} presents the main contribution of the paper: tight convergence guarantees for $\CGD$, $\EF$, and $\EFtw$, along with matching lower bounds.
    Section~\ref{sec:methodology} details the methodology we use to derive the results,
    and provides references to the formal results required to justify this approach. It also
    contains a number of numerical results that illustrate the equivalence between $\EF$ and $\EFtw$,
    and performance characteristics of the three methods.
    Section~\ref{sec:conclusion} summarizes the results of the paper and  
    provides a discussion of the results in relation to the points brought up in the introductory
    section.

\paragraph{Notations:}  We denote $\SO^\ell$ the symmetric matrices, and denote $\SO^\ell_+$ the set of positive semi definite matrices. For any two matrices $A\in \SO^\ell$ and $B\in \SO^d$, we denote $A\otimes B$ the Kronecker product.

\section{Background}\label{sec:background}
In this section, we briefly overview relevant existing results from the field of distributed optimization,
the necessary background on the performance estimation framework, provide the rest of the assumptions we
will need, and specify the notion of Lyapunov functions used in this paper.

\subsection{Theoretical results on  \(\CGD\), \(\EF\), \(\EFtw\)}
In the single agent case, we leverage the equivalence between compressed gradient descent ($\CGD$), under \Cref{as:compression} and the \textit{inexact gradient method} with relatively bounded gradients. $\CGD$ corresponds to the particular case of~\Cref{alg:cgd} with $n=1$, and relatively bounded gradients means that for any $x$, the oracle queried at point $x$ outputs a value $g$ such that $\|g-\nabla f(x)\|^2\le  \epsilon \|\nabla f(x)\|^2$. Various notions of gradient approximation have been studied~\cite{d2008smooth, devolder2014first, schmidt2011}, and a tight analysis for relatively bounded gradients was given in~\cite{de2020worst}.
Specifically, authors have shown that the inexact gradient method then
enjoys tight convergence guarantees for any step size \(\eta > 0\), with respect to the functional 
residual, Euclidean norm distance to the solution and gradient norm.
However, $\CGD$ is known to diverge when applied using stochastic gradient oracles, and to
non-smooth functions~\cite{karimireddy_error_2019}. Interestingly, it is also known to
diverge in the multi-worker setting~\cite{beznosikov_biased_2020}. 
Studying  $\CGD$ is important in its own right, because when the compression operator is chosen 
as the sign function, and the algorithm is applied in the stochastic setting (i.e., signSGD), 
there is a connection to Adam both in the convex~\cite{balles2018dissecting}, and non-convex 
setting~\cite{bernstein_signsgd_2018}. 

Convergence rates for $\EF$ have been established with strongly convex~\cite{stich_sparsified_2018}, quasi-convex and non-convex~\cite{karimireddy_error_2019, stich_error-feedback_2020} functions, and using stochastic gradients. \citet{richtarik_ef21_2021} 
study $\EFtw$ in the multi-worker setting, with (potentially) randomized compression operators. They
establish a \(\UB(k^{-1})\) convergence rate on Lipschitz smooth functions, and a linear rate under
the additional assumption that the functions satisfy the Łojasiewicz inequality. These results 
are obtained using a Lyapunov function; however, without tightness guarantees---neither for the
choice of Lyapunov function, nor for the convergence rate itself. Extensions of $\EFtw$ have
been proposed, including adaptations to stochastic gradients~\cite{fatkhullin_ef21_2021},
and the introduction of a momentum term to improve sample 
complexity in the stochastic setting~\cite{fatkhullin2023momentum}.

\subsection{Performance estimation}
Performance estimation tools~\cite{drori2014performance,taylor2017smooth,taylor2017exact} enable to obtain tight (i.e., exact worst-case) numerical guarantees on convergence rates for various choices of Lyapunov functions. To do so, the estimation of the worst-case rate is formulated as a semidefinite program (SDP), which is then solved numerically using standard solvers such as MOSEK~\cite{mosek}. The resulting numerical values approximate the exact \emph{worst-case} rate of an algorithm over a class of functions, and should not be confused with quantities that depend on specific data, initial points, or problem instances.

This framework has been made accessible through software packages in both Python~\cite{goujaud2022pepit} and Matlab~\cite{taylor2017performance}, enabling researchers to easily apply these tools.
Advanced performance estimation techniques based on the dual formulation of the aforementioned SDP have been developed within this framework to construct optimal Lyapunov functions for first-order methods~\cite{taylor2018lyapunov, upadhyaya2023automated, taylor2019stochastic}.
Particularly relevant is the approach of~\cite{taylor2018lyapunov}, which formulates the search for \emph{quadratic} Lyapunov functions as a feasibility problem with  a candidate contraction rate. By performing bisection on this rate, the method identifies the smallest contraction rate for which a valid Lyapunov function exists.

Another relevant line of work we leverage to discover the analytical form of the Lyapunov functions lies in the field of \emph{symbolic regression}, which
aims to solve supervised learning tasks over the space 
of simple analytic expressions. Recent advances use genetic programming
to search this space, and software packages have been developed in both Python and Julia~\cite{cranmerInterpretableMachineLearning2023}.

\subsection{Definitions \& Notation}
We have already introduced the notion of a contractive compression operator in~\Cref{as:compression}.
To position our contribution within the broader literature we now specify that our
analysis is restricted to the setting of smooth, strongly convex functions:
\begin{assumption}\label{as:smooth}
    The function \(f\) is \(L\)-smooth, i.e., for all \(x, y \in \mathbb{R}^d\), we have
    \[
        f(y) \leq f(x) + \langle \nabla f(x), y - x \rangle + \frac{L}{2} \|y - x\|^2.
    \]
\end{assumption}

\begin{assumption}\label{as:strong_cvx}
    The function \(f\) is \(\mu\)-strongly convex, i.e., for all \(x, y \in \mathbb{R}^d\), we have
      \[
      f(y) \geq f(x) + \langle \nabla f(x), y - x\rangle + \frac{\mu}{2} \|y - x\|^2.
      \]
\end{assumption}
We will  use the notation \(\Fml \) to denote the set of smooth, 
strongly convex functions with parameters~\(\mu\) and \(L\). We will denote \(\kappa \coloneqq \frac{L}{\mu}\) the condition number. For any   objective function \(f\in \Fml\), we  denote $x_\star\coloneqq \arg\min_{x \in \mathbb{R}^d} f$ its minimizer, and \(f_\star \coloneqq \min_{x \in \mathbb{R}^d} f(x)\) its minimum value.

\paragraph{Lyapunov functions.} We now formally define the  class of Lyapunov 
functions under consideration. 

We formally denote \(\M: \R^{\ell \times d} \times \R^d \times \mathcal F \to \R^{\ell \times d} \times \R^d \)  a first-order method acting on a set of functions $\mathcal F$, for an integer $\ell \in \mathbb N$. Such a method, given a function $f\in \mathcal F$,  is applied to an initial \textit{state} \(\xi_0 \in \R^{\ell \times d}\) and iterate \(x_0 \in \R^d\), and generates a sequence
\(\{\xi_k\}_{k \geq 0}\) of states, and a sequence \(\{x_k\}_{k \geq 0}\) of iterations. 
The \textit{states} represent information summarizing the current point in the optimization trajectory that the algorithms may depend on beyond the current iterate---for example, error-related quantities in error feedback algorithms.
The integer $\ell$  is thus typically small, from 0 to 3 in general. The specific states used in this paper are all specified in \Cref{sec:numerical}.

\begin{definition}[Candidate Lyapunov function]\label{def:candidate_lyapunov}
A function \(\Vbase: \mathbb{R}^{\ell \times d} \times  \mathbb{R}^d \to \mathbb{R}\) is called a 
candidate Lyapunov function for \(f\) if it satisfies the following conditions:
\begin{enumerate}[leftmargin=*]
    \item \label{item:non_negativity} (Non-negativity) \(\Lyap{\xi}{x}{f} \geq 0\), for any \(\xi \in \mathbb{R}^{\ell \times d}\), \(x \in \mathbb{R}^d\),
    \item \label{item:zero_fixed_point} (Zero at fixed-point) \(\Lyap{\xi}{x}{f} = 0\) if and only if {\(x = x_\star\)} and \(\xi = \xi_\star\) 
    for a unique \(\xi_\star \in \mathbb{R}^{\ell \times d}\).
    \item \label{item:lower_bound} (Meaningfully lower bounded) 
    there exists a positive semidefinite matrix \(A \in \SO^\ell_+\) and a scalar \(a \geq 0\) such that
    \(\Lyap{\xi}{x}{f} \geq (\xi - \xi_\star)^\top 
    (A \otimes I_d) (\xi - \xi_\star) + a (f(x) - f_\star)\) 
    and \(\tr(A) + a = 1\).
\end{enumerate}
\end{definition}

The lower bound in item~\ref{item:lower_bound} of our definition requires some 
justification: it ensures that the Lyapunov function provides control over 
meaningful quantities in optimization, such as the distance to the fixed point,
gradient norm, algorithm-dependent quantities and the functional residual. 

The class of candidate Lyapunov functions is thus given by
\begin{equation}\label{eq:lyapunov_class_set}
\VCf = \left\{ \Pp \in \SO_{+}^\ell  \times \R^+ :   \tr(P) + p = 1 \right\}.
\end{equation}
For any $\Pp \in \VCf$
we denote $\Vbase_{\Pp}$ the Lyapunov functions  of the form:
\begin{equation}\label{eq:lyapunov_class_form}
    \Lyappp{\xi}{x}{f} = (\xi - \xi_\star)^\top (P \otimes I_d) (\xi - \xi_\star) + p (f(x) - f_\star).
\end{equation}

We seek candidate Lyapunov functions \(\Vbase: \R^{\ell \times d} \times \R^d \times \F \to \R\) that satisfy the  recurrence
\begin{equation}\label{eq:lyapunov_recurrence}
    \Lyap{\xi_{k+1}}{x_{k+1}}{f} \leq \rho \cdot \Lyap{\xi_{k}}{x_{k}}{f},
\end{equation}
for some constant \(\rho < 1\) and for all \(k \geq 0\), uniformly over $\mathcal F$. Finding
the \emph{optimal} Lyapunov function within a parameterized class, for a method $\M$, then amounts to 
solving the following problem:
\begin{equation}\label{eq:lyapunov_problem}
  \rhostar{\M} := \min_{\Pp \in \VCf} \left\{ \max_{\substack{f \in \F_{\mu, L} ,\\ (\xi_0, x_0)\in \mathbb{R}^{\ell \times d} \times  \mathbb{R}^d}} \frac{  \Lyappp{\xi_1}{x_1}{f}}{  \Lyappp{\xi_0}{x_0}{f}} : (\xi_{1}, x_{1}) = \Mfxxi{\xi_0}{x_0}{f} \right\}.
\end{equation}
Note that we set $k=0$ in order to have a guarantee that is valid for all $x \in \R^d$.

\section{Main results} \label{sec:main_results}
In this section, we provide answers to the questions stated in the introduction for our setting. 
We begin by showing some numerical results on the performance of each method, and then provide
the precise statements of all of our theoretical results. 

\subsection{Numerical performance of all methods}
\label{sec:numerical}
In order to compare $\EF$ and $\EFtw$ with the performance of $\CGD$,
we first need to specify the state-variables under consideration when analyzing each method. Those are given
given by: 
\begin{equation}\label{eq:states}
    \xi^{\text{ $\CGD$}}_k = \begin{bmatrix} x_k \\ \Df(x_k) \\ \C(\eta \Df(x_k)) \end{bmatrix}, \quad
    \xi^{\text{EF}}_k = \begin{bmatrix} x_k \\ \Df(x_k) \\ \C(e_k + \eta \Df(x_k)) \\ e_k \end{bmatrix}, \quad
    \xi^{\text{ $\EFtw$}}_k = \begin{bmatrix} x_k \\ \Df(x_k) \\ d_k \end{bmatrix},
\end{equation}
where all variables are defined as in Algorithm~\ref{alg:cgd}, Algorithm~\ref{alg:ef} and Algorithm~\ref{alg:ef21}, respectively. Note that all numerical results of this article are using \emph{deterministic} compression operators.

\begin{figure}[b]
    \centering
    \includegraphics[width=\textwidth]{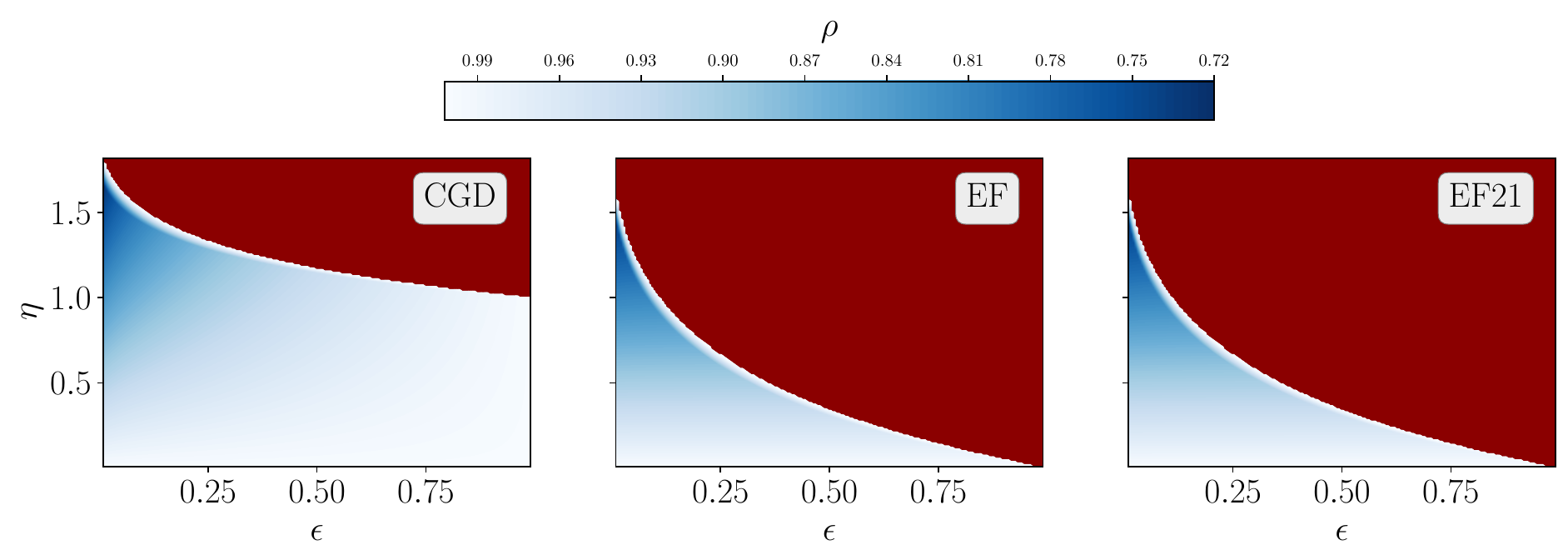}
    \caption{Single row of contour plots showing performance of  $\CGD$, $\EF$, and $\EFtw$ as a function 
    of step size \(\eta\) and compression parameter \(\epsilon\), with regions of non-convergence 
    marked in red. The regions of non-convergence were computed using PEPit by finding
    cycles of length 2.}
    \label{fig:performance_comparison}
\end{figure}

We are now ready to present the numerical results on the performance of each method. 
Figure~\ref{fig:performance_comparison} shows contour plots of the contraction factor for each method.
That is, for a fine grid of both the step size $\eta$ in \Cref{alg:cgd,alg:ef,alg:ef21}, and the parameter $\epsilon$ in  Assumption~\ref{as:compression}, we numerically compute the value of the best possible worst-case contraction rate in terms of our class of candidate Lyapunov functions, over $\Fml$, as given by \eqref{eq:lyapunov_problem}. A darker blue point indicates a stronger contraction $\rhostar{\M}$ (i.e., a better rate). A red point indicates that the method is non-convergent for that choice of $(\epsilon, \eta)$.  
We observe that, although the results are purely numerical at this stage, they indicate that $\EF$ and $\EFtw$ exhibit identical performance in our setting.
This numerical equivalence  is supported by Table~\ref{tab:ef_vs_ef21}: the maximum absolute difference between contraction factors for $\EF$ and $\EFtw$ is on the order of $10^{-5}$ to $10^{-7}$. 
This first fact is a very surprising observation.  Indeed while $\EF$ and $\EFtw$ are known to be 
identical in the very specific case of using a deterministic positively homogeneous and additive compression operator~\cite[see Section 4.2]{richtarik_ef21_2021}, $\EF$ and $\EFtw$ remain grounded in fundamentally different motivations at first sight: on the one hand, $\EF$ accounts for the errors introduced by the compression step, while on the other hand, $\EFtw$ subtracts a control variate from the gradient prior to compression. Proving that the best possible convergence rate they can obtain, for any tuning $\epsilon, \eta$, is similar (but achieved for a \textit{different} Lyapunov function) was, to the best of our knowledge, never established in the literature. It thus constitutes a significant step towards better understanding their connections.

\begin{table}[t]
    \centering
    \begin{minipage}[t]{0.48\textwidth}
        \centering
        \begin{tabular}{lrrr}
\toprule
 & $\kappa = 2$ & $\kappa = 4$ & $\kappa = 10$ \\
\midrule
Absolute error & 4.34e-07 & 6.02e-07 & 1.21e-06 \\
\bottomrule
\end{tabular}

        \caption{Maximum absolute difference of contraction factor for $\EF$ 
        and $\EFtw$, computed over grid of \(\epsilon \in [0.01, 0.99]\) and \(\eta \in [0.01, \frac{2}{L + \mu}]\)
        for \(L = 1\), and varying \(\mu\).}
        \label{tab:ef_vs_ef21}
    \end{minipage} \hfill   \begin{minipage}[t]{0.48\textwidth}
        \centering
        \begin{tabular}{lrrr}
\toprule
 & $\kappa = 2$ & $\kappa = 4$ & $\kappa = 10$ \\
\midrule
Absolute error & 1.14e-06 & 3.90e-07 & 5.11e-06 \\
\bottomrule
\end{tabular}

        \caption{Maximum absolute difference of contraction factor for  $\CGD$ 
        when allowing any combination of terms in our Lyapunov function, 
        compared to the contraction achieved by the functional residual. 
        Same grid as \Cref{tab:ef_vs_ef21}. }
        \label{fig:cgd_optimal_lyapunov}
    \end{minipage}
\end{table}

A second observation can be made from those plots:  the region of non-convergence is by far larger for $\EF$ and $\EFtw$ than for  $\CGD$. In particular, there exist multiple tunings, for which incorporating any of the two types of error feedback, actually \emph{prevent} convergence.

While given for a single $\ml$ in Figure~\ref{fig:performance_comparison} and \Cref{tab:ef_vs_ef21}, similar results hold for all values of $\ml$ that were tried numerically, and several examples are given in Appendix~\ref{app:additional_numerical_results}, along with details of the numerical experiments, including the computation of regions of non-convergence.

Furthermore, we \textit{tune} each algorithm
by picking the optimal step size for each method. We compute the rate $\inf_{\eta} \rhostar{\M_\eta}$, for $\M \in \{\CGD, \EF, \EFtw\}$, where $\M_\eta$ corresponds to the method with step size $\eta$. Results are shown in \Cref{sec:rate_comparison} for three values of $\kappa$, namely 2, 4, and 10. In each setup, for every level of compression, $\CGD$ achieves a rate which is strictly better than $\EF$ and $\EFtw$. In \Cref{sec:comp_cgd_ef}, we also provide a symbolic certifiate of this fact.
These results challenge the prevailing intuition that error feedback ensures convergence comparable to that of uncompressed methods, and even demonstrate that in the single agent and deterministic gradient regime, error feedback is actually always detrimental to convergence.

Finally, we note that the functional residual constitutes an optimal Lyapunov function for $\CGD$,
as shown in Table~\ref{fig:cgd_optimal_lyapunov}. This result is not particularly surprising, given that a tight analysis of the functional residual with the optimal step size was previously established for \textit{inexact gradient descent} by \citet{de2020worst}. That work also demonstrated the tightness of this rate\footnote{Following the same line of reasoning as in our remark on the tightness of our Lyapunov functions in Section~\ref{sec:conclusion}, we can then show that the functional residual is an optimal Lyapunov function for  $\CGD$.}. For $\EFtw$, \citet{richtarik_ef21_2021} proposed another Lyapunov function than the one used here. In \Cref{sec:lyapunov_compare}, \Cref{fig:lyapunov_comparison} we compare the complexities of an optimally tuned version of their Lyapunov function with the class-optimal Lyapunov function numerically. Further comparison of rates can be found in \Cref{sec:rate_comparison}. In \Cref{sec:analytical_comp_richtarik}, we prove that our rate is strictly faster than the the aforementioned rate for $\EFtw$.


In the next two sections, we provide analytical results on $\EF$ and $\EFtw$, respectively.

\subsection{Exact convergence rate and optimal tuning for $\EF$} \label{sec:ef}
We begin by stating the main result of this section, which is a tight rate of
convergence for $\EF$.
\begin{certified}[cas=https://github.com/DanielBergThomsen/error-feedback-tight/blob/main/certificates/EF/upper_bound_CAS.wls, 
    pep=https://github.com/DanielBergThomsen/error-feedback-tight/blob/main/certificates/EF/theorem1_pepit.ipynb, 
    badgeheight=1.5em]{theorem}{theoremEF}\label{thm:ef}
    Consider running Algorithm~\ref{alg:ef}, i.e., $\EF$,  with a compression 
    operator $\mathcal C$ satisfying Assumption~\ref{as:compression} for some \(\epsilon \in [0, 1]\) 
    on any function satisfying Assumptions~\ref{as:smooth}, and~\ref{as:strong_cvx}.
    Let the step size be given by
    \begin{equation}\label{eq:ef_step_size} 
    \eta^\star = \left( \frac{2}{L + \mu} \right) \cdot \left( \frac{1 - \sqrt{\epsilon}}{1 + \sqrt{\epsilon}} \right). 
    \end{equation}
    Then, we have that
    \begin{equation}\label{eq:ef_recurrence}
  \textstyle     \rhostar{\EF_{\eta^\star}}= \sqrt{\epsilon} + \frac{1}{4} (1 + \sqrt{\epsilon}) (L - \mu) \lambda,
    \end{equation}
    where 
    \begin{equation}\label{eq:lagrange_ef}
     \textstyle        \lambda \coloneqq \frac{\eta^\star}{L+\mu} \left[(1 - \sqrt{\epsilon})(L - \mu) + (1 + \sqrt{\epsilon}) \sqrt{(L - \mu)^2 + 16 L \mu \frac{\sqrt{\epsilon}}{(1 + \sqrt{\epsilon})^2}} \right].
    \end{equation}
    A Lyapunov function achieving the rate in~\eqref{eq:ef_recurrence},    with $\xi^{\EF}$ defined in \eqref{eq:states}, is given by
    \begin{equation}\label{eq:lyap_EF}
        \Lyap{\xi^{\EF}}{x}{f} \coloneqq \|x - \xs\|^2 - 2 (x - \xs)^\top e + \left(1 + \frac{1}{\sqrt{\epsilon}}\right) \cdot \|e\|^2 = \|x - \xs - e\|^2 + \frac{1}{\sqrt{\epsilon}} \|e\|^2,
    \end{equation}
    
    Finally, the step size in~\eqref{eq:ef_step_size} is worst-case optimal for $\EF$:  $\forall \eta\geq 0$, we have $ \rhostar{\EF_\eta} \geq  \rhostar{\EF_{\eta^*}}$ .
\end{certified}
Importantly, \eqref{eq:ef_recurrence} shows that the rate is tight; that is, there exist $f \in \Fml$ and $\Pp \in \VCf$ for which the rate is exactly achieved. Since the lower bound also applies to any other performance measure in our state space, our performance measure is optimal. This is formally demonstrated in the proof provided in \Cref{sec:proof_ef}. The proof was written using \emph{deterministic} compressors, but the same guarantees hold under expectation with stochastic compressors as is explained in~\Cref{sec:stoch_extension}.

For completeness—and to support both our theoretical results and the tightness of the numerical results obtained through performance estimation—we provide additional figures comparing the empirically observed optimal step sizes for worst-case instances with our theoretical step size $\eta^\star$, across different values of $\epsilon$ and $\mu/L$. These results are shown in \Cref{fig:ef_step_size}, located in~\Cref{sec:step_size}. The numerical and analytical values match up to numerical accuracy.

\subsection{Exact convergence rate and optimal tuning for $\EFtw$} \label{sec:ef21}
We now state our main result on the $\EFtw$ algorithm, which is also tight.
\begin{certified}[cas=https://github.com/DanielBergThomsen/error-feedback-tight/blob/main/certificates/EF21/upper_bound_CAS.wls, 
    pep=https://github.com/DanielBergThomsen/error-feedback-tight/blob/main/certificates/EF21/theorem2_pepit.ipynb, 
    badgeheight=1.5em]{theorem}{theoremEFtw}\label{thm:ef21}
    Consider running Algorithm~\ref{alg:ef21} with a compression 
    operator satisfying Assumption~\ref{as:compression} for some \(\epsilon \in [0, 1]\) 
    on any function satisfying Assumptions~\ref{as:smooth}, and~\ref{as:strong_cvx}.
Let the step size be given by $\eta^\star$ in~\eqref{eq:ef_step_size}.
    Then, 
\begin{equation}\label{eq:ef21_recurrence}
        \rhostar{\EFtw_{\eta_\star}} = \rhostar{\EF_{\eta_\star}}.
    \end{equation}
    A Lyapunov function achieving the rate in~\eqref{eq:ef21_recurrence} is given by
    \begin{equation}\label{eq:ef21_lyapunov}
        \Lyap{\xi^{\EFtw}}{x}{f} \coloneqq (1 + \sqrt{\epsilon}) \cdot \|g\|^2 - 2 g^\top d + \|d\|^2 = \|g - d\|^2 +  \sqrt{\epsilon} \cdot \|g\|^2.
    \end{equation}
    Finally, the step size $\eta^\star$ is worst-case optimal for this algorithm.
\end{certified}

\begin{remark}
The explicit rate for $\EF$ and $\EFtw$ can be written as
\begin{equation}\label{eq:rate_explicit}
\rho = 
\sqrt{\epsilon} + \left(\frac{1 - \sqrt{\epsilon}}{2}\right) \left(\frac{\kappa - 1}{\kappa + 1}\right)^2 \left[1 - \sqrt{\epsilon} + \sqrt{(1 + \sqrt{\epsilon})^2 + \sqrt{\epsilon} 16 \frac{\kappa}{(\kappa - 1)^2}}\right],
\end{equation}
where $\kappa = L / \mu$. A detailed comparison between this and the existing rate for $\EFtw$ under the Łojasiewicz inequality~\cite{richtarik_ef21_2021} is provided in \Cref{sec:rate_comparison}.
\end{remark}
The proof of this theorem is given in Appendix~\ref{sec:proof_ef21} and assumes \emph{deterministic} compressors, but the same guarantees hold under expectation with stochastic compressors, as is explained in~\Cref{sec:stoch_extension}.

This second theoretical result analytically confirms the surprising numerical observation from \Cref{sec:numerical}, illustrated in \Cref{fig:performance_comparison}:
$\EF$ and $\EFtw$ have the exact same optimal guarantee. Furthermore, the optimal step size is also the same, and can also be argued to be worst-case 
optimal in our setting using the same arguments as in Section~\ref{sec:ef}. However, the optimal Lyapunov function is very different. 

\subsection{Tightness over multiple iterations and choice of state variables}\label{sec:tightness}
A natural question arising from the above analysis concerns the tightness of the Lyapunov functions provided in~\Cref{thm:ef,thm:ef21} \textit{over multiple steps}. Specifically, for $K \geq 2$, we investigate whether the convergence rate of a method $\M \in {\EF, \EFtw, \CGD}$ can be improved by analyzing $\M^K$, the method run over $K$ iterations, defined as follows:
\begin{equation*}
  \rhostarK{\M} := \Bigg(\min_{\Pp \in \VCf} \Bigg\{ \max_{\substack{f \in \F_{\mu, L} ,\\ (\xi_0, x_0)\in \mathbb{R}^{\ell \times d} \times  \mathbb{R}^d}} \frac{  \Lyappp{\xi_K}{x_K}{f}}{  \Lyappp{\xi_0}{x_0}{f}} : (\xi_{K}, x_{K}) = \MfxxiK{\xi_0}{x_0}{f} \Bigg\}\Bigg)^{1/K} \ .
\end{equation*}

We provide a numerical answer to that question in 
Appendix~\ref{sec:lyapunov_multistep} for all algorithms discussed in the paper, showing that the analysis of the single-state Lyapunov functions used in this work are tight
even if we consider multiple iterations. 
To that end, we plot the worst-case contractions for multiple iterations computed using PEPit~\cite{goujaud2022pepit}, on the optimal Lyapunov functions given by \eqref{eq:lyap_EF} and \eqref{eq:ef21_lyapunov}.


\textbf{Lyapunov Tightness.} To prove that the Lyapunov functions we use in Theorems~\ref{thm:ef} 
and~\ref{thm:ef21} are tight, one has to show that the rate of
convergence for any other candidate Lyapunov function from our class is lower bounded by the rate we obtain. We consider a  quadratic function, used in the latter sections of our proofs to prove tightness: asymptotic expressions for all the state-variables are the same, up to an iteration-independent constant.
Consequently, when  computing ratios between any set of Lyapunov functions, these constants cancel out. The only thing remaining is the term that  actually depends the iteration count, which is exactly equal to the rate given in Theorems~\ref{thm:ef} and~\ref{thm:ef21}.

\section{Methodology}\label{sec:methodology}
We now present the methodology used to obtain the results in 
Section~\ref{sec:main_results}. The approach builds on the framework developed by~\cite{taylor2018lyapunov}. We extend it to cover methods using \emph{deterministic} compression operators under Assumption~\ref{as:compression} in
Appendix~\ref{sec:feasibility_compression}. Obtaining the proofs of \Cref{thm:ef,thm:ef21} required a combination of advanced performance estimation techniques (finding optimal Lyapunov functions), several tricks, as well as  symbolic computation and symbolic regression frameworks. 

To solve problem~\eqref{eq:lyapunov_problem}, we begin by addressing
the inner maximization problem for a fixed contraction factor \(\rho\). This amounts to
checking the feasibility of a semidefinite program, detailed in Appendix~\ref{sec:feasibility_compression}.
We then apply bisection on \(\rho\) to identify the smallest admissible 
contraction factor. However, the Lyapunov function given is rarely unique, and most solutions obtained numerically vary significantly with problem parameters such
as the compression factor \(\epsilon\). To address this, we use rank minimization heuristics---specifically, the logdet heuristic~\cite{fazel2003log}. This enables one to obtain a unique set of structurally simpler, low-rank Lyapunov functions. 
Finally, we proceed by eliminating redundant coefficients in the matrix \(P\) and the scalar \(p\), to arrive at the concise forms presented in Section~\ref{sec:main_results}.
At the end of this process, the Lyapunov function coefficients were found to be mutually dependent,
reducing the problem to identifying a closed-form expression for any one of them.
To estimate such a coefficient, we applied symbolic regression using the PySR Python package \cite{cranmerInterpretableMachineLearning2023}. This approach proved highly effective at finding simple yet \emph{optimal} Lyapunov functions. To arrive at simple and readable proofs, we leverage the computer algebra system of \emph{Mathematica}~\cite{Mathematica}.

We wish to emphasize that the combined use of log-det heuristics, symbolic regression, and a computer algebra system turned out to be highly effective at solving this problem, and we believe it has broad applicability to other problems 
in machine learning.

\section{Conclusion}\label{sec:conclusion}
In this paper, we provided tight analyses of $\EF$ and $\EFtw$ using
Lyapunov functions, with guarantees on both the Lyapunov functions themselves and the convergence
rates achieved. Notably, both algorithms exhibit the 
same convergence rate in our setting, and through a remark made in the discussion below,
this gives us tight rates for any of the candidate Lyapunov functions we considered in our class.
We also observed that their performance is strictly worse than that of compressed gradient descent---an 
outcome that, we believe, challenges the intuition of many in the field. 

Our analysis is confined to the single-agent setting, both as an interesting problem on its own, and as a source of intuition for the multi-agent case. $\CGD$
cannot serve as a baseline in the multi-agent setting as it fails to converge with more than one
agent~\cite{beznosikov_biased_2020}. In contrast, $\EFtw$ was specifically designed to \emph{improve} convergence
over $\EF$ in the multi-agent setting. Yet, its convergence rate in the single-agent
case matches exactly that of $\EF$. The findings of this paper raise two compelling questions:
\begin{itemize}[noitemsep,topsep=0pt]
    \item \emph{Does the performance of $\EF$ and $\EFtw$ differ in the multi-agent setting?}
    \item \emph{Are there more effective error compensation mechanisms yet to be discovered?}
\end{itemize}
We leave these questions for future work and conclude by emphasizing that the methodology used
is likely applicable to a broad range of problems in optimization for machine learning.
We look forward to seeing it extended and applied in future research.

\acksection
 We would like to thank Si Yi Meng, Jean-Baptiste Fest, Abel Douzal, and Lucas Versini for providing helpful feedback on an early draft of this paper.
 D.~Berg Thomsen and  A.~Taylor are supported by the European Union (ERC grant CASPER 101162889). The work of A. Dieuleveut is partly supported by ANR-19-CHIA-0002-01/chaire SCAI, and Hi!Paris FLAG project, PEPR Redeem. The French government also partly funded this work under the management of Agence Nationale de la Recherche as part of the ``France 2030'' program, references ANR-23-IACL-0008 "PR[AI]RIE-PSAI", ANR-23-PEIA-005 (REDEEM project) and ANR-23-IACL-0005.

\newpage

\bibliographystyle{unsrtnat}
\bibliography{bib.bib}

\begin{thebibliography}{66}
\providecommand{\natexlab}[1]{#1}
\providecommand{\url}[1]{\texttt{#1}}
\expandafter\ifx\csname urlstyle\endcsname\relax
  \providecommand{\doi}[1]{doi: #1}\else
  \providecommand{\doi}{doi: \begingroup \urlstyle{rm}\Url}\fi

\bibitem[McMahan et~al.(2017)McMahan, Moore, Ramage, Hampson, and Arcas]{mcmahan_communication-efficient_2017}
Brendan McMahan, Eider Moore, Daniel Ramage, Seth Hampson, and Blaise Aguera~y Arcas.
\newblock Communication-{Efficient} {Learning} of {Deep} {Networks} from {Decentralized} {Data}.
\newblock In \emph{{International} {Conference} on {Artificial} {Intelligence} and {Statistics} {(AISTATS)}}, April 2017.

\bibitem[Kairouz et~al.(2019)Kairouz, McMahan, Avent, Bellet, Bennis, Bhagoji, Bonawitz, Charles, Cormode, Cummings, D'Oliveira, Rouayheb, Evans, Gardner, Garrett, Gascón, Ghazi, Gibbons, Gruteser, Harchaoui, He, He, Huo, Hutchinson, Hsu, Jaggi, Javidi, Joshi, Khodak, Konečný, Korolova, Koushanfar, Koyejo, Lepoint, Liu, Mittal, Mohri, Nock, Özgür, Pagh, Raykova, Qi, Ramage, Raskar, Song, Song, Stich, Sun, Suresh, Tramèr, Vepakomma, Wang, Xiong, Xu, Yang, Yu, Yu, and Zhao]{kairouz_advances_2019}
Peter Kairouz, H.~Brendan McMahan, Brendan Avent, Aurélien Bellet, Mehdi Bennis, Arjun~Nitin Bhagoji, Keith Bonawitz, Zachary Charles, Graham Cormode, Rachel Cummings, Rafael G.~L. D'Oliveira, Salim~El Rouayheb, David Evans, Josh Gardner, Zachary Garrett, Adrià Gascón, Badih Ghazi, Phillip~B. Gibbons, Marco Gruteser, Zaid Harchaoui, Chaoyang He, Lie He, Zhouyuan Huo, Ben Hutchinson, Justin Hsu, Martin Jaggi, Tara Javidi, Gauri Joshi, Mikhail Khodak, Jakub Konečný, Aleksandra Korolova, Farinaz Koushanfar, Sanmi Koyejo, Tancrède Lepoint, Yang Liu, Prateek Mittal, Mehryar Mohri, Richard Nock, Ayfer Özgür, Rasmus Pagh, Mariana Raykova, Hang Qi, Daniel Ramage, Ramesh Raskar, Dawn Song, Weikang Song, Sebastian~U. Stich, Ziteng Sun, Ananda~Theertha Suresh, Florian Tramèr, Praneeth Vepakomma, Jianyu Wang, Li~Xiong, Zheng Xu, Qiang Yang, Felix~X. Yu, Han Yu, and Sen Zhao.
\newblock Advances and {Open} {Problems} in {Federated} {Learning}.
\newblock \emph{arXiv:1912.04977 [cs, stat]}, December 2019.

\bibitem[Seide et~al.(2014)Seide, Fu, Droppo, Li, and Yu]{seide_1-bit_2014}
Frank Seide, Hao Fu, Jasha Droppo, Gang Li, and Dong Yu.
\newblock 1-{Bit} {Stochastic} {Gradient} {Descent} and its {Application} to {Data}-{Parallel} {Distributed} {Training} of {Speech} {DNNs}.
\newblock In \emph{{Annual} {Conference} of the {International} {Speech} {Communication} {Association}}, 2014.

\bibitem[Chilimbi et~al.(2014)Chilimbi, Suzue, Apacible, and Kalyanaraman]{chilimbi_project_2014}
Trishul Chilimbi, Yutaka Suzue, Johnson Apacible, and Karthik Kalyanaraman.
\newblock Project adam: Building an efficient and scalable deep learning training system.
\newblock In \emph{{USENIX} {Symposium} on {Operating} {Systems} {Design} and {Implementation} ({OSDI} 14)}, 2014.

\bibitem[Strom(2015)]{strom_scalable_2015}
Nikko Strom.
\newblock Scalable distributed {DNN} training using commodity {GPU} cloud computing.
\newblock In \emph{{Annual} {Conference} of the {International} {Speech} {Communication} {Association}}, 2015.

\bibitem[Karimireddy et~al.(2020)Karimireddy, Kale, Mohri, Reddi, Stich, and Suresh]{karimireddy_scaffold_2020}
Sai~Praneeth Karimireddy, Satyen Kale, Mehryar Mohri, Sashank Reddi, Sebastian Stich, and Ananda~Theertha Suresh.
\newblock Scaffold: Stochastic controlled averaging for federated learning.
\newblock In \emph{International Conference on Machine Learning (ICML)}, 2020.

\bibitem[Alistarh et~al.(2017)Alistarh, Grubic, Li, Tomioka, and Vojnovic]{alistarh_qsgd_2017}
Dan Alistarh, Demjan Grubic, Jerry Li, Ryota Tomioka, and Milan Vojnovic.
\newblock {QSGD}: {Communication}-{Efficient} {SGD} via {Gradient} {Quantization} and {Encoding}.
\newblock \emph{Advances in Neural Information Processing Systems (NeurIPS)}, 30, 2017.

\bibitem[Alistarh et~al.(2018)Alistarh, Hoefler, Johansson, Konstantinov, Khirirat, and Renggli]{alistarh_convergence_2018}
Dan Alistarh, Torsten Hoefler, Mikael Johansson, Nikola Konstantinov, Sarit Khirirat, and Cedric Renggli.
\newblock The {Convergence} of {Sparsified} {Gradient} {Methods}.
\newblock \emph{Advances in Neural Information Processing Systems (NeurIPS)}, 31, 2018.

\bibitem[Mishchenko et~al.(2024)Mishchenko, Gorbunov, Takáč, and and]{mishchenko_distributed_2019}
K.~Mishchenko, E.~Gorbunov, M.~Takáč, and P.~Richtárik and.
\newblock Distributed learning with compressed gradient differences.
\newblock \emph{Optimization Methods and Software}, 2024.

\bibitem[Karimireddy et~al.(2019)Karimireddy, Rebjock, Stich, and Jaggi]{karimireddy_error_2019}
Sai~Praneeth Karimireddy, Quentin Rebjock, Sebastian Stich, and Martin Jaggi.
\newblock Error {Feedback} {Fixes} {SignSGD} and other {Gradient} {Compression} {Schemes}.
\newblock In \emph{International {Conference} on {Machine} {Learning} ({ICML})}, 2019.

\bibitem[Wu et~al.(2018)Wu, Huang, Huang, and Zhang]{wu_error_2018}
Jiaxiang Wu, Weidong Huang, Junzhou Huang, and Tong Zhang.
\newblock Error {Compensated} {Quantized} {SGD} and its {Applications} to {Large}-scale {Distributed} {Optimization}.
\newblock In \emph{International {Conference} on {Machine} {Learning} ({ICML})}, 2018.

\bibitem[Horvath et~al.(2022)Horvath, Ho, Horvath, Sahu, Canini, and Richt{\'a}rik]{horvath_natural_2022}
Samuel Horvath, Chen-Yu Ho, Ludovit Horvath, Atal~Narayan Sahu, Marco Canini, and Peter Richt{\'a}rik.
\newblock {Natural} {Compression} for {Distributed} {Deep} {Learning}.
\newblock In \emph{Mathematical and Scientific Machine Learning}, 2022.

\bibitem[Li et~al.(2020)Li, Kovalev, Qian, and Richtarik]{li_acceleration_2020}
Zhize Li, Dmitry Kovalev, Xun Qian, and Peter Richtarik.
\newblock Acceleration for {Compressed} {Gradient} {Descent} in {Distributed} and {Federated} {Optimization}.
\newblock In \emph{International {Conference} on {Machine} {Learning} ({ICML})}, 2020.

\bibitem[Richt\'arik et~al.(2021)Richt\'arik, Sokolov, and Fatkhullin]{richtarik_ef21_2021}
Peter Richt\'arik, Igor Sokolov, and Ilyas Fatkhullin.
\newblock {EF21}: {A} {New}, {Simpler}, {Theoretically} {Better}, and {Practically} {Faster} {Error} {Feedback}.
\newblock In \emph{Advances in {Neural} {Information} {Processing} {Systems} ({NeurIPS})}, 2021.

\bibitem[Harrane et~al.(2018)Harrane, Flamary, and Richard]{harrane_reducing_2018}
Ibrahim El~Khalil Harrane, R{\'e}mi Flamary, and C{\'e}dric Richard.
\newblock On reducing the communication cost of the diffusion lms algorithm.
\newblock \emph{IEEE Transactions on Signal and Information Processing over Networks}, 5\penalty0 (1):\penalty0 100--112, 2018.

\bibitem[Tang et~al.(2019)Tang, Yu, Lian, Zhang, and Liu]{tang_doublesqueeze_2019}
Hanlin Tang, Chen Yu, Xiangru Lian, Tong Zhang, and Ji~Liu.
\newblock {DoubleSqueeze}: {Parallel} {Stochastic} {Gradient} {Descent} with {Double}-pass {Error}-{Compensated} {Compression}.
\newblock In \emph{International {Conference} on {Machine} {Learning} ({ICML})}, 2019.

\bibitem[Liu et~al.(2020)Liu, Li, Tang, and Yan]{liu_double_2020}
Xiaorui Liu, Yao Li, Jiliang Tang, and Ming Yan.
\newblock A {Double} {Residual} {Compression} {Algorithm} for {Efficient} {Distributed} {Learning}.
\newblock In \emph{International {Conference} on {Artificial} {Intelligence} and {Statistics} (AISTATS)}, 2020.

\bibitem[Zheng et~al.(2019)Zheng, Huang, and Kwok]{zheng_communication-efficient_2019}
Shuai Zheng, Ziyue Huang, and James Kwok.
\newblock Communication-{Efficient} {Distributed} {Blockwise} {Momentum} {SGD} with {Error}-{Feedback}.
\newblock In \emph{Advances in {Neural} {Information} {Processing} {Systems} ({NeurIPS})}, 2019.

\bibitem[Philippenko and Dieuleveut(2020)]{philippenko_artemis_2020}
Constantin Philippenko and Aymeric Dieuleveut.
\newblock Bidirectional compression in heterogeneous settings for distributed or federated learning with partial participation: tight convergence guarantees.
\newblock \emph{arXiv:2006.14591 [cs, stat]}, 2020.

\bibitem[Philippenko and Dieuleveut(2021)]{philippenko_preserved_2021}
Constantin Philippenko and Aymeric Dieuleveut.
\newblock Preserved central model for faster bidirectional compression in distributed settings.
\newblock \emph{{Advances} in {Neural} {Information} {Processing} {Systems} ({NeurIPS})}, 2021.

\bibitem[Gorbunov et~al.(2020{\natexlab{a}})Gorbunov, Kovalev, Makarenko, and Richtarik]{gorbunov_linearly_2020}
Eduard Gorbunov, Dmitry Kovalev, Dmitry Makarenko, and Peter Richtarik.
\newblock Linearly {Converging} {Error} {Compensated} {SGD}.
\newblock In \emph{Advances in {Neural} {Information} {Processing} {Systems} ({NeurIPS})}, 2020{\natexlab{a}}.

\bibitem[Sattler et~al.(2019)Sattler, Wiedemann, Müller, and Samek]{sattler_robust_2019}
Felix Sattler, Simon Wiedemann, Klaus-Robert Müller, and Wojciech Samek.
\newblock Robust and {Communication}-{Efficient} {Federated} {Learning} {From} {Non}-i.i.d. {Data}.
\newblock \emph{IEEE Transactions on Neural Networks and Learning Systems}, pages 3400--3413, 2019.
\newblock ISSN 2162-2388.
\newblock \doi{10.1109/TNNLS.2019.2944481}.

\bibitem[Fatkhullin et~al.(2021)Fatkhullin, Sokolov, Gorbunov, Li, and Richtárik]{fatkhullin_ef21_2021}
Ilyas Fatkhullin, Igor Sokolov, Eduard Gorbunov, Zhize Li, and Peter Richtárik.
\newblock {EF21} with {Bells} \& {Whistles}: {Practical} {Algorithmic} {Extensions} of {Modern} {Error} {Feedback}, October 2021.
\newblock arXiv:2110.03294 [cs, math].

\bibitem[Beznosikov et~al.(2023)Beznosikov, Horváth, Richtárik, and Safaryan]{beznosikov_biased_2020}
Aleksandr Beznosikov, Samuel Horváth, Peter Richtárik, and Mher Safaryan.
\newblock On {Biased} {Compression} for {Distributed} {Learning}.
\newblock \emph{Journal of Machine Learning Research}, 24\penalty0 (276):\penalty0 1--50, 2023.

\bibitem[Chraibi et~al.(2019)Chraibi, Khaled, Kovalev, Richt{\'a}rik, Salim, and Tak{\'a}{\v{c}}]{chraibi_distributed_2019}
S{\'e}lim Chraibi, Ahmed Khaled, Dmitry Kovalev, Peter Richt{\'a}rik, Adil Salim, and Martin Tak{\'a}{\v{c}}.
\newblock {Distributed} {Fixed} {Point} {Methods} with {Compressed} {Iterates}.
\newblock \emph{arXiv:1912.09925 [cs, math]}, December 2019.

\bibitem[Gorbunov et~al.(2020{\natexlab{b}})Gorbunov, Hanzely, and Richt{\'a}rik]{gorbunov_unified_2020}
Eduard Gorbunov, Filip Hanzely, and Peter Richt{\'a}rik.
\newblock A {Unified} {Theory} of {SGD}: {Variance} {Reduction}, {Sampling}, {Quantization} and {Coordinate} {Descent}.
\newblock In \emph{International Conference on Artificial Intelligence and Statistics (AISTATS)}, 2020{\natexlab{b}}.

\bibitem[Reisizadeh et~al.(2020)Reisizadeh, Mokhtari, Hassani, Jadbabaie, and Pedarsani]{reisizadeh_fedpaq_2020}
Amirhossein Reisizadeh, Aryan Mokhtari, Hamed Hassani, Ali Jadbabaie, and Ramtin Pedarsani.
\newblock {FedPAQ}: {A} {Communication}-{Efficient} {Federated} {Learning} {Method} with {Periodic} {Averaging} and {Quantization}.
\newblock In \emph{International {Conference} on {Artificial} {Intelligence} and {Statistics} (AISTATS)}, 2020.

\bibitem[Kovalev et~al.(2021)Kovalev, Gasanov, Gasnikov, and Richtarik]{kovalev_lower_2021}
Dmitry Kovalev, Elnur Gasanov, Alexander Gasnikov, and Peter Richtarik.
\newblock {Lower} {Bounds} and {Optimal} {Algorithms} for {Smooth} and {Strongly} {Convex} {Decentralized} {Optimization} {Over} {Time}-{Varying} {Networks}.
\newblock \emph{Advances in Neural Information Processing Systems (NeurIPS)}, 2021.

\bibitem[Haddadpour et~al.(2021)Haddadpour, Kamani, Mokhtari, and Mahdavi]{haddadpour_federated_2021}
Farzin Haddadpour, Mohammad~Mahdi Kamani, Aryan Mokhtari, and Mehrdad Mahdavi.
\newblock {Federated} {Learning} with {Compression}: {Unified} {Analysis} and {Sharp} {Guarantees}.
\newblock In \emph{International Conference on Artificial Intelligence and Statistics (AISTATS)}, 2021.

\bibitem[Li and Richt{\'a}rik(2021)]{li_canita_2021}
Zhize Li and Peter Richt{\'a}rik.
\newblock {CANITA}: {Faster} {Rates} for {Distributed} {Convex} {Optimization} with {Communication} {Compression}.
\newblock \emph{Advances in Neural Information Processing Systems (NeurIPS)}, 2021.

\bibitem[Khirirat et~al.(2018)Khirirat, Feyzmahdavian, and Johansson]{khirirat_distributed_2018}
Sarit Khirirat, Hamid~Reza Feyzmahdavian, and Mikael Johansson.
\newblock Distributed learning with compressed gradients.
\newblock \emph{arXiv:1806.06573 [cs, math]}, June 2018.

\bibitem[Stich et~al.(2018)Stich, Cordonnier, and Jaggi]{stich_sparsified_2018}
Sebastian~U Stich, Jean-Baptiste Cordonnier, and Martin Jaggi.
\newblock Sparsified {SGD} with {Memory}.
\newblock In \emph{Advances in {Neural} {Information} {Processing} {Systems} ({NeurIPS})}. 2018.

\bibitem[Ivkin et~al.(2019)Ivkin, Rothschild, Ullah, Braverman, Stoica, and Arora]{ivkin_communication-efficient_2019}
Nikita Ivkin, Daniel Rothschild, Enayat Ullah, Vladimir Braverman, Ion Stoica, and Raman Arora.
\newblock Communication-efficient {Distributed} {SGD} with {Sketching}.
\newblock \emph{Advances in Neural Information Processing Systems (NeurIPS)}, 2019.

\bibitem[Koloskova et~al.(2019)Koloskova, Stich, and Jaggi]{koloskova_decentralized_2019}
Anastasia Koloskova, Sebastian Stich, and Martin Jaggi.
\newblock Decentralized {Stochastic} {Optimization} and {Gossip} {Algorithms} with {Compressed} {Communication}.
\newblock In \emph{International {Conference} on {Machine} {Learning} ({ICML})}, 2019.

\bibitem[Stich and Karimireddy(2020)]{stich_error-feedback_2020}
Sebastian~U Stich and Sai~Praneeth Karimireddy.
\newblock The error-feedback framework: {Better} rates for {SGD} with delayed gradients and compressed updates.
\newblock \emph{Journal of Machine Learning Research}, 21:\penalty0 1--36, 2020.

\bibitem[Fatkhullin et~al.(2023)Fatkhullin, Tyurin, and Richt{\'a}rik]{fatkhullin2023momentum}
Ilyas Fatkhullin, Alexander Tyurin, and Peter Richt{\'a}rik.
\newblock Momentum provably improves error feedback!
\newblock \emph{Advances in Neural Information Processing Systems (NeurIPS)}, 2023.

\bibitem[Gruntkowska et~al.(2023)Gruntkowska, Tyurin, and Richt\'{a}rik]{gruntkowska_ef21-p_2022}
Kaja Gruntkowska, Alexander Tyurin, and Peter Richt\'{a}rik.
\newblock {EF}21-{P} and {Friends}: {Improved} {Theoretical} {Communication} {Complexity} for {Distributed} {Optimization} with {Bidirectional} {Compression}.
\newblock In \emph{International Conference on Machine Learning (ICML)}, 2023.

\bibitem[Makarenko et~al.(2022)Makarenko, Gasanov, Islamov, Sadiev, and Richt{\'a}rik]{makarenko2022adaptive}
Dmitry Makarenko, Elnur Gasanov, Rustem Islamov, Abdurakhmon Sadiev, and Peter Richt{\'a}rik.
\newblock {Adaptive} {Compression} for {Communication}-{Efficient} {Distributed} {Training}.
\newblock \emph{arXiv:2211.00188 [cs]}, October 2022.

\bibitem[Richtarik et~al.(2022)Richtarik, Sokolov, Gasanov, Fatkhullin, Li, and Gorbunov]{richtarik2022threepoint}
Peter Richtarik, Igor Sokolov, Elnur Gasanov, Ilyas Fatkhullin, Zhize Li, and Eduard Gorbunov.
\newblock 3{PC}: {Three} {Point} {Compressors} for {Communication}-{Efficient} {Distributed} {Training} and a {Better} {Theory} for {Lazy} {Aggregation}.
\newblock In \emph{International Conference on Machine Learning (ICML)}, 2022.

\bibitem[Zhao et~al.(2022)Zhao, Li, Li, Richt{\'a}rik, and Chi]{zhao2022beer}
Haoyu Zhao, Boyue Li, Zhize Li, Peter Richt{\'a}rik, and Yuejie Chi.
\newblock {BEER}: {Fast} {$O(1/T)$} {Rate} for {Decentralized} {Nonconvex} {Optimization} with {Communication} {Compression}.
\newblock In \emph{Advances in Neural Information Processing Systems (NeurIPS)}, 2022.

\bibitem[Wang et~al.(2022)Wang, Lin, and Chen]{wang2022communication}
Yujun Wang, Lu~Lin, and Jinghui Chen.
\newblock {Communication}-{Compressed} {Adaptive} {Gradient} {Method} for {Distributed} {Nonconvex} {Optimization}.
\newblock In \emph{International {Conference} on Artificial Intelligence and Statistics (AISTATS)}, 2022.

\bibitem[Dorfman et~al.(2023)Dorfman, Vargaftik, Ben-Itzhak, and Levy]{dorfman2023docofl}
Ron Dorfman, Shay Vargaftik, Yaniv Ben-Itzhak, and Kfir~Yehuda Levy.
\newblock {D}o{C}o{FL}: Downlink compression for cross-device federated learning.
\newblock In \emph{International Conference on Machine Learning (ICML)}, 2023.

\bibitem[Lin et~al.(2020)Lin, Stich, Barba, Dmitriev, and Jaggi]{lin2020dynamic}
Tao Lin, Sebastian~U. Stich, Luis Barba, Daniil Dmitriev, and Martin Jaggi.
\newblock Dynamic model pruning with feedback.
\newblock In \emph{International Conference on Learning Representations (ICLR)}, 2020.

\bibitem[Drori(2014)]{drori2014contributions}
Yoel Drori.
\newblock \emph{Contributions to the Complexity Analysis of Optimization Algorithms}.
\newblock PhD thesis, Tel-Aviv University, 2014.

\bibitem[Taylor et~al.(2017{\natexlab{a}})Taylor, Hendrickx, and Glineur]{taylor2017exact}
Adrien~B. Taylor, Julien~M. Hendrickx, and Fran{\c{c}}ois Glineur.
\newblock {Exact} {Worst}-{Case} {Performance} of {First}-{Order} {Methods} for {Composite} {Convex} {Optimization}.
\newblock \emph{SIAM Journal on Optimization}, 27\penalty0 (3):\penalty0 1283--1313, 2017{\natexlab{a}}.

\bibitem[Taylor et~al.(2018)Taylor, Van~Scoy, and Lessard]{taylor2018lyapunov}
Adrien Taylor, Bryan Van~Scoy, and Laurent Lessard.
\newblock {Lyapunov} {Functions} for {First}-{Order} {Methods}: {Tight} {Automated} {Convergence} {Guarantees}.
\newblock In \emph{International Conference on Machine Learning (ICML)}, 2018.

\bibitem[Upadhyaya et~al.(2025)Upadhyaya, Banert, Taylor, and Giselsson]{upadhyaya2023automated}
Manu Upadhyaya, Sebastian Banert, Adrien~B. Taylor, and Pontus Giselsson.
\newblock {Automated} tight {Lyapunov} analysis for first-order methods.
\newblock \emph{Mathematical Programming}, 209\penalty0 (1):\penalty0 133--170, 2025.

\bibitem[Goujaud et~al.(2023{\natexlab{a}})Goujaud, Dieuleveut, and Taylor]{goujaud2023fundamental}
Baptiste Goujaud, Aymeric Dieuleveut, and Adrien Taylor.
\newblock {On} {Fundamental} {Proof} {Structures} in {First}-{Order} {Optimization}.
\newblock In \emph{Conference on Decision and Control (CDC)}, 2023{\natexlab{a}}.

\bibitem[d'Aspremont(2008)]{d2008smooth}
Alexandre d'Aspremont.
\newblock {Smooth} {Optimization} with {Approximate} {Gradient}.
\newblock \emph{SIAM Journal on Optimization}, 19\penalty0 (3):\penalty0 1171--1183, 2008.

\bibitem[Devolder et~al.(2014)Devolder, Glineur, and Nesterov]{devolder2014first}
Olivier Devolder, Fran{\c{c}}ois Glineur, and Yurii Nesterov.
\newblock First-order methods of smooth convex optimization with inexact oracle.
\newblock \emph{Mathematical Programming}, 146:\penalty0 37--75, 2014.

\bibitem[Schmidt et~al.(2011)Schmidt, Roux, and Bach]{schmidt2011}
Mark Schmidt, Nicolas Roux, and Francis Bach.
\newblock {Convergence} {Rates} of {Inexact} {Proximal}-{Gradient} {Methods} for {Convex} {Optimization}.
\newblock In \emph{Advances in Neural Information Processing Systems (NeurIPS)}, 2011.

\bibitem[De~Klerk et~al.(2020)De~Klerk, Glineur, and Taylor]{de2020worst}
Etienne De~Klerk, Francois Glineur, and Adrien~B Taylor.
\newblock {Worst}-{Case} {Convergence} {Analysis} of {Inexact} {Gradient} and {Newton} {Methods} {Through} {Semidefinite} {Programming} {Performance} {Estimation}.
\newblock \emph{SIAM Journal on Optimization}, 30\penalty0 (3):\penalty0 2053--2082, 2020.

\bibitem[Balles and Hennig(2018)]{balles2018dissecting}
Lukas Balles and Philipp Hennig.
\newblock {Dissecting} {Adam}: {The} {Sign}, {Magnitude} and {Variance} of {Stochastic} {Gradients}.
\newblock In \emph{International Conference on Machine Learning (ICML)}, 2018.

\bibitem[Bernstein et~al.(2018)Bernstein, Wang, Azizzadenesheli, and Anandkumar]{bernstein_signsgd_2018}
Jeremy Bernstein, Yu-Xiang Wang, Kamyar Azizzadenesheli, and Animashree Anandkumar.
\newblock {signSGD}: {Compressed} {Optimisation} for {Non}-{Convex} {Problems}.
\newblock In \emph{International Conference on Machine Learning (ICML)}, 2018.

\bibitem[Drori and Teboulle(2014)]{drori2014performance}
Yoel Drori and Marc Teboulle.
\newblock Performance of first-order methods for smooth convex minimization: a novel approach.
\newblock \emph{Mathematical Programming}, 145\penalty0 (1):\penalty0 451--482, 2014.

\bibitem[Taylor et~al.(2017{\natexlab{b}})Taylor, Hendrickx, and Glineur]{taylor2017smooth}
Adrien~B. Taylor, Julien~M. Hendrickx, and Fran{\c{c}}ois Glineur.
\newblock Smooth strongly convex interpolation and exact worst-case performance of first-order methods.
\newblock \emph{Mathematical Programming}, 161\penalty0 (1-2):\penalty0 307--345, 2017{\natexlab{b}}.

\bibitem[ApS(2025)]{mosek}
MOSEK ApS.
\newblock \emph{MOSEK Optimizer API for Python 11.0.21}, 2025.
\newblock URL \url{https://docs.mosek.com/11.0/pythonapi/index.html}.

\bibitem[Goujaud et~al.(2024)Goujaud, Moucer, Glineur, Hendrickx, Taylor, and Dieuleveut]{goujaud2022pepit}
Baptiste Goujaud, C{\'e}line Moucer, Fran{\c c}ois Glineur, Julien~M. Hendrickx, Adrien~B. Taylor, and Aymeric Dieuleveut.
\newblock {PEPit}: computer-assisted worst-case analyses of first-order optimization methods in {Python}.
\newblock \emph{Mathematical Programming Computation}, 16\penalty0 (3):\penalty0 337--367, 2024.

\bibitem[Taylor et~al.(2017{\natexlab{c}})Taylor, Hendrickx, and Glineur]{taylor2017performance}
Adrien~B. Taylor, Julien~M. Hendrickx, and Fran{\c{c}}ois Glineur.
\newblock {Performance estimation toolbox (PESTO): automated worst-case analysis of first-order optimization methods}.
\newblock In \emph{Conference on Decision and Control (CDC)}, 2017{\natexlab{c}}.

\bibitem[Taylor and Bach(2019)]{taylor2019stochastic}
Adrien Taylor and Francis Bach.
\newblock Stochastic first-order methods: non-asymptotic and computer-aided analyses via potential functions.
\newblock In \emph{Conference on Learning Theory (COLT)}, 2019.

\bibitem[Cranmer(2023)]{cranmerInterpretableMachineLearning2023}
Miles Cranmer.
\newblock Interpretable {Machine} {Learning} for {Science} with {PySR} and {SymbolicRegression}.jl.
\newblock \emph{arXiv:2305.01582 [physics]}, May 2023.

\bibitem[Fazel et~al.(2003)Fazel, Hindi, and Boyd]{fazel2003log}
Maryam Fazel, Haitham Hindi, and Stephen~P. Boyd.
\newblock Log-det heuristic for matrix rank minimization with applications to {Hankel} and {Euclidean} distance matrices.
\newblock In \emph{American Control Conference (ACC)}, 2003.

\bibitem[Inc.()]{Mathematica}
Wolfram~Research{,} Inc.
\newblock Mathematica, {V}ersion 14.2.
\newblock URL \url{https://www.wolfram.com/mathematica}.
\newblock Champaign, IL, 2024.

\bibitem[Gannot(2022)]{gannot2022frequency}
Oran Gannot.
\newblock A frequency-domain analysis of inexact gradient methods.
\newblock \emph{Mathematical Programming}, 194\penalty0 (1-2):\penalty0 975--1016, 2022.

\bibitem[De~Klerk et~al.(2017)De~Klerk, Glineur, and Taylor]{de2017worst}
Etienne De~Klerk, Fran{\c{c}}ois Glineur, and Adrien~B Taylor.
\newblock On the worst-case complexity of the gradient method with exact line search for smooth strongly convex functions.
\newblock \emph{Optimization Letters}, 11:\penalty0 1185--1199, 2017.

\bibitem[Goujaud et~al.(2023{\natexlab{b}})Goujaud, Dieuleveut, and Taylor]{goujaud2023counter}
Baptiste Goujaud, Aymeric Dieuleveut, and Adrien Taylor.
\newblock Counter-examples in first-order optimization: a constructive approach.
\newblock \emph{IEEE Control Systems Letters}, 2023{\natexlab{b}}.
\newblock {(See \href{https://arxiv.org/abs/2303.10503}{arXiv 2303 10503} for complete version with appendices)}.

\end{thebibliography}

\clearpage

\appendix
\section*{Organization of the appendix}
\newcommand{\customtoc}[1]{
    \item[\ref{#1}] \nameref{#1} \dotfill \pageref{#1}
}
\begin{itemize}[leftmargin=2em]
    \customtoc{sec:feasibility_compression}
    \customtoc{sec:rate_comparison}
    \customtoc{sec:proofs}
    \customtoc{app:additional_numerical_results}
    \customtoc{sec:proof_certs}
\end{itemize}

This appendix provides additional content and details complementing the paper. In particular, \Cref{sec:feasibility_compression} details the general methodology used to search for Lyapunov functions. The complete missing proofs for the main results of the paper are presented in~\Cref{sec:proofs}.  \Cref{app:additional_numerical_results} presents additional numerical results that informally motivate a few choices made in the paper and provide numerical validation of our claims.
Finally, \Cref{sec:proof_certs} discusses our choice of annotating this article with proof certificates.

\section{Feasibility problems with compressors} \label{sec:feasibility_compression}
This section presents the methodology used to search for Lyapunov functions. In a nutshell, we formulate the Lyapunov search problem as a quasi-convex optimization problem involving linear matrix inequalities. Those problems are typically solved through the use of an iterative procedure involving a binary search with semidefinite solvers---we use MOSEK~\cite{mosek} throughout. 
The main steps taken here can be viewed as a generalization of the procedure proposed in~\cite{taylor2018lyapunov} to first-order methods using compression, in particular \Cref{alg:ef,alg:ef21}. We also simplify a few steps that are not needed for our purposes.
We start by reviewing the technique on a simpler example in~\Cref{a:baby_GD} before detailing the more tricky formulations involving compression.

\subsection{Feasibility problem for gradient descent}\label{a:baby_GD}
To introduce the concepts underlying the techniques used to construct Lyapunov functions for \Cref{alg:ef,alg:ef21}, we begin with the simpler case of \emph{gradient descent} on smooth, strongly convex functions.
That is, we consider the algorithm:
\begin{equation}\label{eq:GDeta}
    x_1=x_0-\eta \nabla f(x_0) \tag{$\mathrm{GD}_\eta$}. 
\end{equation}
The goal of this subsection is to review the steps used to compute $\rho^\star(\mathrm{GN}_\eta)$, as defined in \eqref{eq:lyapunov_problem}, via a bisection search in which each iteration involves verifying the feasibility of a convex problem.

Our starting point is to consider the following state variable for GD:
\begin{equation}\label{eq:states_gd}
    \xi^{\text{$\mathrm{GD}$}}_k = \begin{bmatrix} x_k \\ \Df(x_k) \end{bmatrix}
\end{equation}
and a natural family of Lyapunov function candidates (which corresponds to a subset of \eqref{eq:lyapunov_class_set}) of the form
\begin{equation}
\begin{aligned}
    \mathcal{V}_P(\xi^{\text{$\mathrm{GD}$}}_k, x_k ; f) \equiv \mathcal{V}_P(x_k, \nabla f(x_k) ; f) &\coloneqq 
    \begin{bmatrix} x_k - x_\star \\ \nabla f(x_k) \end{bmatrix}^\top
    (P\otimes I_d)
    \begin{bmatrix} x_k - x_\star \\ \nabla f(x_k) \end{bmatrix}\\&=P_{11}\|x_k-x_\star\|^2+P_{22}\|\nabla f(x_k)\|^2\\&+2P_{12}\langle \nabla f(x_k);x_k-x_\star\rangle,
\end{aligned}
\end{equation}
where \(P \in \SO^d\) is positive semidefinite and we require \(\tr(P) = 1\). This latter requirement is without loss of generality due to a normalization argument, and is added to avoid the trivial solution $P=0$.

The problem we aim to solve is that of finding the best Lyapunov function among a given set of candidates—specifically, the one for which the ratio \( \frac{\mathcal{V}_P(x_1, \nabla f(x_1))}{\mathcal{V}_P(x_0, \nabla f(x_0))} \) can be uniformly upper bounded by the smallest possible constant over all optimization problems in the considered family. In other words, we seek the Lyapunov candidate function that yields the smallest possible $\rho$ such that
\begin{equation}\label{eq:LyapPraterho}
 \frac{\mathcal{V}_P(x_1, \nabla f(x_1); f)}{\mathcal{V}_P(x_0, \nabla f(x_0); f)} \leq \rho
\end{equation}
is valid for all $L$-smooth $\mu$-strongly convex functions $f:\mathbb{R}^d\rightarrow \mathbb{R}$ (in any dimension $d\in\mathbb{N}$) and all possible $x_0,x_1,x_\star \in\mathbb{R}^d$ compatible with $f$ and $x_1=x_0-\eta \nabla f(x_0)$. The problem can be phrased as finding 
\begin{equation}
    \begin{aligned}
 \rho^\star(\mathrm{GD}_\eta) =    \min_{\substack{ P \succeq 0}} \bigg(
    \max_{\substack{d\in\mathbb{N}\\f \in \Fml\\ x_0,x_\star \in \R^d}} \left\{\frac{\mathcal{V}_P(x_1, \nabla f(x_1); f)}{\mathcal{V}_P(x_0, \nabla f(x_0); f)}  \quad \text{s.t.} \quad
    \tr(P) = 1,\,x_1 = x_0 - \eta \nabla f(x_0)\right\}\bigg)
    \end{aligned}
\end{equation}
where \(\eta>0\) is the step size. One can reformulate this problem directly using tools from the performance estimation literature~\cite{drori2014performance,taylor2017smooth}, but the resulting problem is not convex in the variable \(\rho\). 
One way to address this is to reduce it to the problem of finding, for a given contraction factor~\(\rho\), some Lyapunov function that achieves this~\(\rho\)---if one exists.
This problem, on the other hand, is convex, and we can simply perform bisection search on $\rho$ to find the smallest possible contraction factor.

We now introduce some notation to simplify the statement of finding such Lyapunov functions:
\begin{equation}\label{eq:sigmaasLyapdiff}
    \sigma_\rho(x_1, g_1, x_0, g_0; P) \coloneqq \mathcal{V}_P(x_1, g_1; f) - \rho \mathcal{V}_P(x_0, g_0; f).
\end{equation}
We will arrive at a way of solving this problem by reasoning through two steps:
\begin{enumerate}
    \item \textbf{Step 1:} verifying a \emph{given} Lyapunov function and rate as a convex problem.
    \item \textbf{Step 2:} verifying a rate $\rho$ for the \emph{optimal} candidate Lyapunov functions as a convex problem.
\end{enumerate}

\paragraph{\textbf{Step 1:} verifying a \emph{given} Lyapunov function and rate as a convex problem.} \ \\
For a fixed Lyapunov parameter $P \in \mathbb V_2$ and a tentative rate $\rho>0$,  we can then state the problem of \emph{verifying} a given Lyapunov function as that of showing that the minimum value of the following is  non-positive:
\begin{equation}\label{eq:lyapPraterhoasSigma}
    \begin{aligned}
0\geq    \sup_{\substack{d\in\mathbb{N}\\ f\in \Fml\\x_0,x_\star \in \R^d }} \quad & \sigma_\rho(x_1, \Df(x_1), x_0, \Df(x_0); P; f) \\
    \text{s.t.} \quad &  x_1 = x_0 - \eta \nabla f(x_0) \\
    &\nabla f(x_\star)=0\\
    \end{aligned}
\end{equation}
The constraint that \(f\) is a smooth strongly convex function is easily encoded using
interpolation conditions \cite{taylor2017smooth}. This allows us to work with 
\emph{sampled points} from \(f\) rather than the infinite dimensional set \(\Fml\).
We introduce the notation
\begin{equation}\label{eq:interp_app}
    \phi_{ij} \coloneqq f_i - f_j - g_j^\top (x_i - x_j) - \frac{1}{2L} \|g_i - g_j\|^2
    - \frac{\mu}{2(1 - \mu/L)} \|x_i - x_j - \frac{1}{L} (g_i - g_j)\|^2,
\end{equation}
where the notation \((x_i, g_i, f_i)\) is used to denote a sampled triplet from \(f\),
such that \(f(x_i) = f_i\) and \(\nabla f(x_i) = g_i\) for all \(i \in \{0, 1, \star\}\).
This lets us rephrase our problem as
\begin{equation}\label{eq:baby_pep}
    \begin{aligned}
    0\geq \sup_{\substack{d\in\mathbb{N}\\x_\star, x_0, g_\star, g_0, g_1 \in \R^d \\ f_0, f_1\in\R}} \quad & \sigma_\rho(x_1, g_1, x_0, g_0; P; f) \\
    \text{s.t.} \quad & \phi_{ij} \geq 0 \qquad \forall i, j \in \{0, 1, \star\} \\
    & x_1 = x_0 - \eta \nabla f(x_0)\\
    & g_\star=0\\
    \end{aligned}
\end{equation}
The above problem is not convex due to the interpolation constraints. To address this, we
reformulate it as a semidefinite program (SDP). Let \(G = B^\top B\) where \(B\) is the
following \((3 \times d)\) matrix
\begin{equation*}
    B = \begin{bmatrix} x_0- x_\star,\, g_0,\, g_1 \end{bmatrix},
\end{equation*}
and let \(\mathbf{f} \coloneqq \begin{bmatrix} f_0-f_\star\\ f_1-f_\star \end{bmatrix}\). In other words, $G\succeq 0$ is the Gram matrix of the entries of $B$:
\[ \begin{bmatrix}
    \|x_0-x_\star\|^2 & \langle g_0,x_0-x_\star\rangle & \langle g_1,x_0-x_\star\rangle \\
    \langle g_0,x_0-x_\star\rangle & \| g_0\|^2 & \langle g_1,g_0\rangle \\
    \langle g_1,x_0-x_\star\rangle & \langle g_1,g_0\rangle & \| g_1\|^2 
\end{bmatrix}\succeq 0.\]

We introduce a convenient notation by defining basis (row) vectors in $\bar{x}_i,\bar{g}_i\in \mathbb{R}^3$ and $\bar{f}_i\in\mathbb{R}^2$ which allow us to ``select'' specific elements in $B$ and $\mathbf{f}$. Specifically, we define them such that
\begin{equation}
 x_i-x_\star=B\bar{x}_i^\top,\, \quad g_i=B\bar{g}_i^\top,\, \quad f_i-f_\star=\bar{f}_i\mathbf{f}.\label{eq:picking}
\end{equation}
More precisely: $\bar{x}_i=\mathbf{e}_1^\top\in\mathbb{R}^3$, $\bar{g}_0=\mathbf{e}_2^\top\in\mathbb{R}^3$, $\bar{g}_1=\mathbf{e}_3^\top\in\mathbb{R}^3$, $\bar{x}_1=\bar{x}_0-\eta \bar{g}_0$, $\bar{x}_\star=\mathbf{0}_3^\top\in\mathbb{R}^3$ along with $\bar{f}_0=\mathbf{e}_1^\top\in\mathbb{R}^2$, $\bar{f}_1=\mathbf{e}_2^\top\in\mathbb{R}^2$, $\bar{f}_\star=\mathbf{0}_2^\top\in\mathbb{R}^2$. We can conveniently rewrite the different parts of problem~\eqref{eq:baby_pep} using the notation
\begin{equation*}
    \begin{aligned}
        \|x_i-x_\star\|^2&=\bar{x}_i B^\top B \bar{x}_i^\top=\tr (\bar{x}_i^\top\bar{x}_i G),\\
        \|g_i\|^2&=\bar{g}_i B^\top B \bar{g}_i^\top=\tr (\bar{g}_i^\top\bar{g}_i G),\\
        \langle g_i,x_j-x_\star\rangle&=\bar{g}_i B^\top B \bar{x}_j^\top=\tr \left( (\bar{g}_i\odot\bar{x}_j) G\right)\\
    \end{aligned}
\end{equation*}
where $\bar{g}_i\odot\bar{x}_j=\tfrac12(\bar{g}_i^\top\bar{x}_j+\bar{x}_j^\top\bar{g}_i)$, and $\odot$ denotes the symmetric outer product. This notation allows us to express all relevant quantities in terms of traces involving symmetric matrices. We now reformulate the necessary terms to derive the desired semidefinite representation of the problem. Let us begin with
\begin{equation*}
    \begin{aligned}
        \mathcal{V}_P(x_0, \nabla f(x_0); f)&= P_{11}  \tr(\bar{x}_0^\top\bar{x}_0 G)+P_{22}  \tr(\bar{g}_0^\top\bar{g}_0 G)+ 2P_{12}  \tr((\bar{x}_0\odot\bar{g}_0) G)\\
        & =\tr(A_0^\top P A_0 G),
    \end{aligned}
\end{equation*}
where $A_0\coloneqq\begin{bmatrix}
    \bar{x}_0\\\bar{g}_0 
\end{bmatrix}$. Similarly, we can write $\mathcal{V}_P(x_1, \nabla f(x_1); f)=\tr(A_1^\top P A_1 G)$ with $A_1\coloneqq\begin{bmatrix}
    \bar{x}_1\\\bar{g}_1 
\end{bmatrix}$ and also define the matrices $M_{ij}$ for all $i,j\in\{0,1,\star\}$ such that $\phi_{ij}=m_{ij}\mathbf{f}+\tr(M_{ij}G)$:
\begin{equation}\label{eq:Mijdef}
    \begin{aligned}
        m_{ij}=&\bar{f}_i-\bar{f}_j\\
    M_{ij}=&-\bar{g}_j\odot(\bar{x}_i-\bar{x}_j) -\tfrac1{2L}(\bar{g}_i-\bar{g}_j)^\top (\bar{g}_i-\bar{g}_j)\\ &\,-\tfrac{\mu}{2(1-\mu/L)}(\bar{x}_i-\bar{x}_j-\tfrac1L(\bar{g}_i-\bar{g}_j))^\top(\bar{x}_i-\bar{x}_j-\tfrac1L(\bar{g}_i-\bar{g}_j))
    \end{aligned}
\end{equation}
This enables a convenient reformulation of~\eqref{eq:baby_pep} as the verification of the following condition, which corresponds to solving a standard semidefinite program:
\begin{equation}\label{eq:Lyaprate_convex}
    \begin{aligned}
   0\geq \sup_{\substack{G \succeq 0\\ \mathbf{f}}} \quad & \tr(A_1^\top P A_1 G) - \rho \tr(A_0^\top P A_0 G) \\
    \text{s.t.} \quad & \tr(M_{ij} G) + \mathbf{f}^\top c_{ij} \geq 0 \qquad \forall i, j \in \{0, 1, \star\}. \\
    \end{aligned}
\end{equation}
One immediate consequence is that the validity of a given Lyapunov function $\mathcal{V}_P$ for a specified rate $\rho$---satisfying \eqref{eq:LyapPraterho} or, equivalently, \eqref{eq:lyapPraterhoasSigma}---can be formulated as the convex problem \eqref{eq:Lyaprate_convex}.

\paragraph{\textbf{Step 2:} verifying a rate $\rho$ for the \emph{optimal} candidate Lyapunov functions as a convex problem.} \ \\
To derive a convenient condition that formally guarantees the above problem is nonpositive, we consider its standard Lagrangian dual\footnote{Strong duality holds in this case due to the existence of a Slater point; see, e.g.,~\cite{taylor2017smooth}.}---which reduces to verifying the existence of dual variables $\lambda_{ij} \geq 0$ such that
\begin{equation}\label{eq:Check_rate_lyap}
    \begin{aligned}
        0\geq \inf_{\lambda_{ij} \geq 0} \quad & 0 \\
        \text{s.t.} \quad & A_1^\top P A_1 - \rho A_0^\top P A_0 - \sum_{i,j\in\{0,1,\star\}} \lambda_{ij} M_{ij} \succeq 0 \\
        & \sum_{i,j\in\{0,1,\star\}} \lambda_{ij} m_{ij} = 0.\\
    \end{aligned}
\end{equation}
The problem has finally reduced to showing that for a given matrix \(P\), the small-sized problem~\eqref{eq:Check_rate_lyap} is feasible. 

\textbf{Key idea:} As this problem is linear in~$P$ (when $\rho$ is fixed), we can directly use it to search for a valid~$P$ that verifies the decrease condition in \eqref{eq:lyapunov_recurrence} for a given $\rho$:
\begin{equation} \tag{GD-SDP} \label{eq:gd_sdp}
    \underset{\substack{P \succeq 0, \\ \lambda_{ij} \geq 0}}{\text{feasible}}\left\{ \quad
    \begin{aligned}
        &  A_1^\top P A_1 - \rho A_0^\top P A_0 - \sum_{i,j} \lambda_{ij} M_{ij} \succeq 0\\
        & \sum_{i,j} \lambda_{ij} m_{ij} = 0\\
        &\tr(P) = 1.
    \end{aligned}\right.
\end{equation}

\paragraph{Conclusion.} 
Problem \eqref{eq:gd_sdp} is a convex feasibility problem that encodes the existence of a candidate Lyapunov function certifying a given rate $\rho$. By performing a bisection search over $\rho$, we can identify the smallest rate satisfied by \emph{some} Lyapunov function in our class. This approach enables us to solve the problem numerically using a semidefinite solver.

\subsection{Feasibility problem for \(\EF\)}\label{a:EF_feas}
Using the same ideas as in \Cref{a:baby_GD}, this section states the feasibility problem we solve to identify Lyapunov functions
for \Cref{alg:ef}. In short, this requires adapting the two steps presented in \Cref{a:baby_GD} to accommodate the compressed message. Practically speaking, \textbf{Step 1} must be adapted to a slightly larger problem (with a few more states both in the Gram matrix $G$ and in the Lyapunov candidates to incorporate compression). 
Then, \textbf{Step 2} follows directly by the same reasoning as before: the condition derived in Step 1 reduces to checking the feasibility of a linear matrix inequality that is linear in $(P, p)$, which in turn allows us to search for the Lyapunov function via binary search over $\rho$.

Recall that we defined our state space for $\EF$ in \eqref{eq:states} as 
\begin{equation*}
    \xi^{\text{EF}}_i = \begin{bmatrix} x_i \\ \Df(x_i) \\ \C(e_i + \eta \Df(x_i)) \\ e_i \end{bmatrix} .
\end{equation*}
This space has dimension 4, and our normalized set of candidate Lyapunov functions is given by
\begin{equation*}
\mathbb V_4 = \left\{ \Pp \in \SO_{+}^4  \times \R^+ :   \tr(P) + p = 1 \right\}.
\end{equation*}
For any $\Pp \in \mathbb V_4$
we thus consider $\Vbase_{\Pp}$ Lyapunov functions of the form:
\begin{equation}\label{eq:lyapunov_shape_EF}
    \Lyappp{\xi^\EF}{x}{f} = (\xi^\EF - \xi^\EF_\star)^\top (P \otimes I_d) (\xi^\EF - \xi^\EF_\star) + p (f(x) - f_\star).
\end{equation}
where we impose $\tr(P)+p=1$, again, without loss of generality and to avoid the trivial solution $(P,p)=0$.
Similarly to \eqref{eq:sigmaasLyapdiff}, we now define
\begin{equation}\label{eq:LyapdiffEF}
    \sigma^\EF_\rho(\xi^\EF_1, \xi^\EF_0; \Pp; f) \coloneqq \mathcal{V}_{\Pp}(\xi^\EF_1, x_1; f) - \rho \mathcal{V}_{\Pp}(\xi^\EF_0, x_0; f).
\end{equation}

Again, we say that for $\Pp \in \mathbb V_4$, a Lyapunov function $\mathcal V_{\Pp}$  \textit{satisfies rate $\rho$}  for the iterates of $\EF$ 
if we have that
\begin{equation}\label{eq:lyapPraterhoasSigmaEF}
    \begin{aligned}
0\geq    \sup_{\substack{d\in\mathbb{N}\\ f\in \Fml\\x_0,x_\star \in \R^d }} \quad & \sigma^\EF_\rho(\xi^\EF_1,\xi^\EF_0; \Pp; f) \\
    \text{s.t.} \quad &  (x_1, \xi^\EF_1) = \EF(x_0 ,\xi^\EF_0; f) \\
    &\nabla f(x_\star)=0\\
    \end{aligned}
\end{equation}
Formally, we require the following lemma:
\begin{lemma}[$\EF$ feasibility problem] Consider running \Cref{alg:ef} with a deterministic compression operator satisfying \Cref{as:compression} for some \(\epsilon \in [0, 1]\) on any function satisfying \Cref{as:smooth,as:strong_cvx}. There exists a nonzero candidate Lyapunov function $\mathcal V_{\Pp}$ of the form defined in \eqref{eq:lyapunov_shape_EF}, satisfying a given rate $\rho>0$, if and only if the following problem is feasible:
    \begin{equation} \tag{\(\EF\)-SDP} \label{eq:ef_sdp}
    \underset{\substack{P \in \SO^{4}, \\ p \in \R, \\ \lambda_{ij} \geq 0, \\ \nu_i \geq 0}}{\text{feasible}} \left\{ \quad
    \begin{aligned}
        &0 \succeq \Delta V_P (\rho) + \sum_{i,j \in \{0, 1, \star\}} \lambda_{ij} M_{ij} + \sum_{i \in \{0, 1\}} \nu_i \cdot C^{\EF}_i \\
        &0 \geq_2 \Delta v_p(\rho) + \sum_{i,j \in \{0, 1, \star\}} \lambda_{ij} m_{ij} \\
        &0 \preceq P \\
        &0 \leq p\\
        & 1= \tr(P)+p
    \end{aligned} \right.
\end{equation}
where matrices $(M_{ij})_{i,j \in \{0, 1, \star\}}$ are defined as in \eqref{eq:Mijdef}, $(C_{i}^{\EF})_{i \in \{0, 1\}}$ given below in \eqref{eq:compression_operator_ef}, and $\Delta V_P(\rho), \Delta v_p(\rho)$ are given below in \eqref{eq:decrease_ef_quadratic} and \eqref{eq:decrease_ef_linear}. Here $\geq_2$ denotes coordinate-wise inequality in $\mathbb R^2$.
\end{lemma}

\begin{proofsketch} 
The proof is decomposed into several steps that correspond to adapting the technical ingredients from \Cref{a:baby_GD}.

\paragraph{Basis vector encoding} 
We begin by introducing notation analogously to \eqref{eq:picking} for $\EF$. Define the following
basis vectors \(\bar{x}_i, \bar{g}_i, \bar{c}_i, \bar{e}_i \in \R^8\):
\begin{equation}
    \bar{x}_i \coloneqq \mathbf{e}_{i+1}^\top, \quad \bar{g}_i \coloneqq \mathbf{e}_{i+3}^\top, \quad \bar{c}_i \coloneqq \mathbf{e}_{i+5}^\top, \quad \bar{e}_i \coloneqq \mathbf{e}_{i+7}^\top,\quad i \in \{0, 1\},
\end{equation}
where  \(\mathbf{e}_i\) is the \(i\)-th basis vector in dimension 8. Similarly, let  \(\bar{f}_i \in \R^2\) be defined by
\begin{equation}
     \bar{f}_i \coloneqq \mathbf{e}_i^\top, \quad i \in \{0, 1\},
\end{equation}
where \(\mathbf{e}_i\) is the \(i\)-th basis vector in dimension 2.

The point of defining these vectors is the same as it was for GD. These vectors allow us to ``select'' points from our Gram matrix (see \eqref{eq:picking}). This, in turn, allows us to express the interpolation conditions of our feasibility problem in a clean manner.

We also define
row vectors that correspond to the fixed-point as:
\begin{align*}
    \bar{x}_\star &\coloneqq \mathbf{0}_8^\top,\quad  
    \bar{g}_\star \coloneqq \mathbf{0}_8^\top,\quad  
    \bar{c}_\star \coloneqq \mathbf{0}_8^\top,\quad  
    \bar{e}_\star \coloneqq \mathbf{0}_8^\top,  \\ 
    \bar{f}_\star &\coloneqq \mathbf{0}_2^\top.
\end{align*}
Finally, we define our method in terms of our basis vectors:
\begin{equation}\label{eq:ef_basis}
    \begin{aligned}
        \bar{x}_1 &= \bar{x}_0 - \bar{c}_0, \\
        \bar{e}_1 &= \bar{e}_0 + \eta \bar{g}_0 - \bar{c}_0.
    \end{aligned}
\end{equation}

\paragraph{Expressing \eqref{eq:LyapdiffEF} using basis vectors.} 

First, we encode the decrease in the linear and quadratic terms of \eqref{eq:LyapdiffEF} as 
\begin{equation}\label{eq:decrease_ef_quadratic}
    \Delta V_P(\rho) \coloneqq 
    \begin{bmatrix} \bar{x}_1 - \bar{x}_\star \\ \bar{g}_1 - \bar{g}_\star \\ \bar{c}_1 - \bar{c}_\star \\ \bar{e}_1 - \bar{e}_\star \end{bmatrix}^\top
    P
    \begin{bmatrix} \bar{x}_1 - \bar{x}_\star \\ \bar{g}_1 - \bar{g}_\star \\ \bar{c}_1 - \bar{c}_\star \\ \bar{e}_1 - \bar{e}_\star \end{bmatrix}
    - \rho \begin{bmatrix} \bar{x}_0 - \bar{x}_\star \\ \bar{g}_0 - \bar{g}_\star \\ \bar{c}_0 - \bar{c}_\star \\ \bar{e}_0 - \bar{e}_\star \end{bmatrix}^\top
    P
    \begin{bmatrix} \bar{x}_0 - \bar{x}_\star \\ \bar{g}_0 - \bar{g}_\star \\ \bar{c}_0 - \bar{c}_\star \\ \bar{e}_0 - \bar{e}_\star \end{bmatrix}, \quad
\end{equation}
\begin{equation}\label{eq:decrease_ef_linear}
    \Delta v_p(\rho) \coloneqq p \left(\bar{f}_1 - \bar{f}_\star \right) 
    - \rho \cdot p \left(\bar{f}_0 - \bar{f}_\star \right),
\end{equation}
where \(\rho > 0\) is the contraction factor to be verified. Note that $\Delta V_P(\rho) \in \mathbb R ^{8\times 8}$, and  $\Delta v_p(\rho) \in \mathbb R^2$.

Using these objects, we have that:
 \begin{equation}
     \sigma^\EF_\rho(\xi^\EF_1, \xi^\EF_0; \Pp; f) =  \tr\Big((\Delta V_P(\rho)) G^\EF\Big) + (\Delta v_p(\rho))^\top F^\EF,
 \end{equation}
where $G^\EF=(B^\EF)^\top B^\EF $ is the Gram matrix of vectors $$B^\EF = \begin{bmatrix} x_0 ,x_1 , \Df(x_0) ,\Df(x_1) , \C(e_0 + \eta \Df(x_0)) ,\C(e_1 + \eta \Df(x_1)) , e_0, e_1 \end{bmatrix}, $$
and $ F^\EF = (f(x_0), f(x_1))$.
 
\paragraph{Interpolation conditions}
The interpolation conditions that enforce $f \in \Fml$ are identical to those we define in \eqref{eq:Mijdef}, but using the new basis vectors we define specifically for $\EF$. We do, however, need to introduce a new interpolation condition to encode the fact that we are using a contractive compressor. We do this for \emph{deterministic} compression operators. Using our basis vectors, this corresponds to introducing the matrices
\begin{equation}\label{eq:compression_operator_ef}
C_i^{\EF} = (\eta \bar{g}_i + \bar{e}_i - \bar{c}_i)^\top (\eta \bar{g}_i + \bar{e}_i - \bar{c}_i) - \epsilon \cdot (\eta \bar{g}_i + \bar{e}_i)^\top (\eta \bar{g}_i + \bar{e}_i),
\end{equation}
for \(i \in \{0, 1\}\).

Finally, following the same reasoning as described in \textbf{Step 1} and \textbf{Step 2} of \Cref{a:baby_GD}, and using the technical modifications we outlined here, we arrive at the feasibility problem described in the statement of the lemma.
\end{proofsketch}

\subsection{Feasibility problem for \(\EFtw\)}\label{a:EF21_feas}
As in the previous section, one can now adapt the same ideas as in \Cref{a:baby_GD} and \Cref{a:EF_feas} to \(\EFtw\). This section states the feasibility problem we solve to identify Lyapunov functions
in this context.

Following a parallel line of derivation, we begin by noting that the state space defined in \eqref{eq:states} for \(\EFtw\) was 
\begin{equation}    
    \xi^{\text{ $\EFtw$}}_k = \begin{bmatrix} x_k \\ \Df(x_k) \\ d_k \end{bmatrix},
\end{equation}

This space now has dimension 3, and our normalized set of candidate Lyapunov functions is given by
\begin{equation*}
\mathbb V_3 = \left\{ \Pp \in \SO_{+}^3  \times \R^+ :   \tr(P) + p = 1 \right\}.
\end{equation*}
For any $\Pp \in \mathbb V_3$
we thus consider $\Vbase_{\Pp}$ Lyapunov functions of the form:
\begin{equation}\label{eq:lyapunov_shape_EF21}
    \Lyappp{\xi^\EFtw}{x}{f} = (\xi^\EFtw - \xi^\EFtw_\star)^\top (P \otimes I_d) (\xi^\EFtw - \xi^\EFtw_\star) + p (f(x) - f_\star).
\end{equation}
where we once more require $\tr(P)+p=1$ without loss of generality and to avoid the trivial solution $(P,p)=0$. Similarly, to \eqref{eq:sigmaasLyapdiff} and \eqref{eq:LyapdiffEF}, we define here
\begin{equation}\label{eq:LyapdiffEF21}
    \sigma^\EFtw_\rho(\xi^\EFtw_1, \xi^\EFtw_0; \Pp; f) \coloneqq \mathcal{V}_{\Pp}(\xi^\EFtw_1, x_1; f) - \rho \mathcal{V}_{\Pp}(\xi^\EFtw_0, x_0; f).
\end{equation}

Again, we say that for $\Pp \in \mathbb V_3$, a Lyapunov $\mathcal V_{\Pp}$  \textit{satisfies rate $\rho$}  for the iterates of $\EFtw$, if we have that
\begin{equation}\label{eq:lyapPraterhoasSigma21}
    \begin{aligned}
0\geq    \sup_{\substack{d\in\mathbb{N}\\ f\in \Fml\\x_0,x_\star \in \R^d }} \quad & \sigma^\EFtw_\rho(\xi^\EFtw_1,\xi^\EFtw_0; \Pp; f) \\
    \text{s.t.} \quad &  (x_1, \xi^\EFtw_1) = \EFtw(x_0 ,\xi^\EFtw_0; f) \\
    &\nabla f(x_\star)=0\\
    \end{aligned}
\end{equation}
Formally, we prove the following lemma.
\begin{lemma}[$\EFtw$ feasibility problem] Consider running \Cref{alg:ef21} with a deterministic compression operator satisfying \Cref{as:compression} for some \(\epsilon \in [0, 1]\) on any function satisfying \Cref{as:smooth,as:strong_cvx}. There exists a nonzero candidate Lyapunov function $\mathcal V_{\Pp}$ of the form defined in \eqref{eq:lyapunov_shape_EF21}, satisfying a given rate $\rho>0$, if and only if the following problem is feasible:
    \begin{equation} \tag{\(\EFtw\)-SDP} \label{eq:ef_sdp21}
    \underset{\substack{P \in \SO^{3 }, \\ p \in \R, \\ \lambda_{ij} \geq 0, \\ \nu_i \geq 0}}{\text{feasible}} \left\{ \quad
    \begin{aligned}
        &0 \succeq \Delta V_P (\rho) + \sum_{i,j \in \{0, 1, \star\}} \lambda_{ij} M_{ij} + \nu \cdot C^{\EFtw}_i \\
        &0 \geq_2 \Delta v_p(\rho) + \sum_{i,j \in \{0, 1, \star\}} \lambda_{ij} m_{ij} \\
        &0 \preceq P \\
        &0 \leq p\\
        & 1= \tr(P)+p
    \end{aligned} \right.
\end{equation}
where matrices $(M_{ij})_{i,j \in \{0, 1, \star\}}$ are defined as in \Cref{eq:Mijdef}, $(C_{i}^{\EFtw})_{i \in \{0, 1\}}$ given below in \eqref{eq:compression_operator_ef21}, and $\Delta V_P(\rho), \Delta v_p(\rho)$ are given below in \eqref{eq:decrease_ef21_quadratic} and \eqref{eq:decrease_ef21_linear}. Here $\geq_2$ denotes coordinate-wise inequality in $\mathbb R^2$.
\end{lemma}
\begin{proofsketch}
We begin by changing our basis vectors to
\begin{equation}
    \bar{x}_i \coloneqq \mathbf{e}_{i+1}^\top, \bar{g}_i \coloneqq \mathbf{e}_{i+3}^\top, \bar{c}_i \coloneqq \mathbf{e}_{i+5}^\top, \bar{d}_i \coloneqq \mathbf{e}_{i+7}^\top, \bar{f}_i \coloneqq \mathbf{e}_i^\top, \quad i \in \{0, 1\}
\end{equation}
where \(\bar{x}_i, \bar{g}_i, \bar{c}_i, \bar{d}_i \in \R^8\), and \(\bar{f}_i \in \R^2\).
Similarly, we define the row vectors corresponding to the fixed-point as
\begin{equation}
    \bar{x}_\star \coloneqq \mathbf{0}_8^\top, \bar{g}_\star \coloneqq \mathbf{0}_8^\top, \bar{c}_\star \coloneqq \mathbf{0}_8^\top, \bar{d}_\star \coloneqq \mathbf{0}_8^\top, \bar{f}_\star \coloneqq \mathbf{0}_2^\top,
\end{equation}
We define our method using these basis vectors as
\begin{equation}
    \begin{aligned}
        \bar{x}_{1} &= \bar{x}_0 + \eta \cdot \bar{d}_0, \\
        \bar{d}_{1} &= \bar{d}_0 + \bar{c}_0.
    \end{aligned}
\end{equation}
The only difference in the interpolation conditions is that we are compressing a different
quantity now, which we can encode using
\begin{equation}\label{eq:compression_operator_ef21}
    C^{\EFtw} \coloneqq 
    (\bar{g}_1 - \bar{d}_0 - \bar{c}_0)^\top (\bar{g}_1 - \bar{d}_0 - \bar{c}_0) - \epsilon (\bar{g}_1 - \bar{d}_0)^\top (\bar{g}_1 - \bar{d}_0).
\end{equation}
where we only need a single matrix because the compression operator acts on a mixture of the
current and next state.

Next, we encode the linear and quadratic terms in exactly the same way as we did for \(\EF\),
except with the state variables
\begin{equation}\label{eq:decrease_ef21_quadratic}
    \Delta V_P(\rho) \coloneqq 
    \begin{bmatrix} \bar{x}_1 - \bar{x}_\star \\ \bar{g}_1 - \bar{g}_\star \\ \bar{d}_1 - \bar{d}_\star \end{bmatrix}^\top
    P
    \begin{bmatrix} \bar{x}_1 - \bar{x}_\star \\ \bar{g}_1 - \bar{g}_\star \\ \bar{d}_1 - \bar{d}_\star \end{bmatrix}
    - \rho \begin{bmatrix} \bar{x}_0 - \bar{x}_\star \\ \bar{g}_0 - \bar{g}_\star \\ \bar{d}_0 - \bar{d}_\star \end{bmatrix}^\top
    P
    \begin{bmatrix} \bar{x}_0 - \bar{x}_\star \\ \bar{g}_0 - \bar{g}_\star \\ \bar{d}_0 - \bar{d}_\star \end{bmatrix}.
\end{equation}
\begin{equation}\label{eq:decrease_ef21_linear}
    \Delta v_p(\rho) \coloneqq p \left(\bar{f}_1 - \bar{f}_\star \right) 
    - \rho \cdot p \left(\bar{f}_0 - \bar{f}_\star \right),
\end{equation}
Here too,  $\Delta V_P(\rho) \in \mathbb R ^{8\times 8}$, and  $\Delta v_p(\rho) \in \mathbb R^2$.

Finally, using \textbf{Step 1} and \textbf{Step 2} of \Cref{a:baby_GD}, we arrive at the feasibility problem described in the statement of the lemma.
\end{proofsketch}

\newpage

\section{Rate comparison}
\label{sec:rate_comparison}
\begin{figure}[h]
    \centering
    \includegraphics[width=\textwidth]{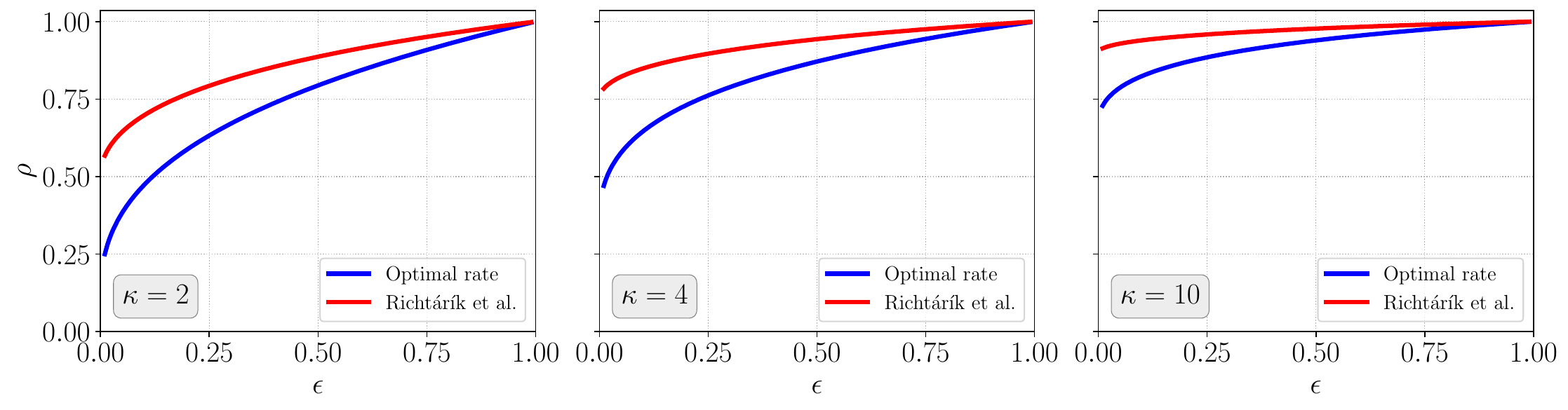}
    \caption{Line plot comparing the convergence rate of this paper (blue) with~\cref{eq:rho_richtarik} (red) for various $\kappa$.}
    \label{fig:rate_comparison_rictharik}
\end{figure}

\begin{figure}[h]
    \centering
    \includegraphics[width=\textwidth]{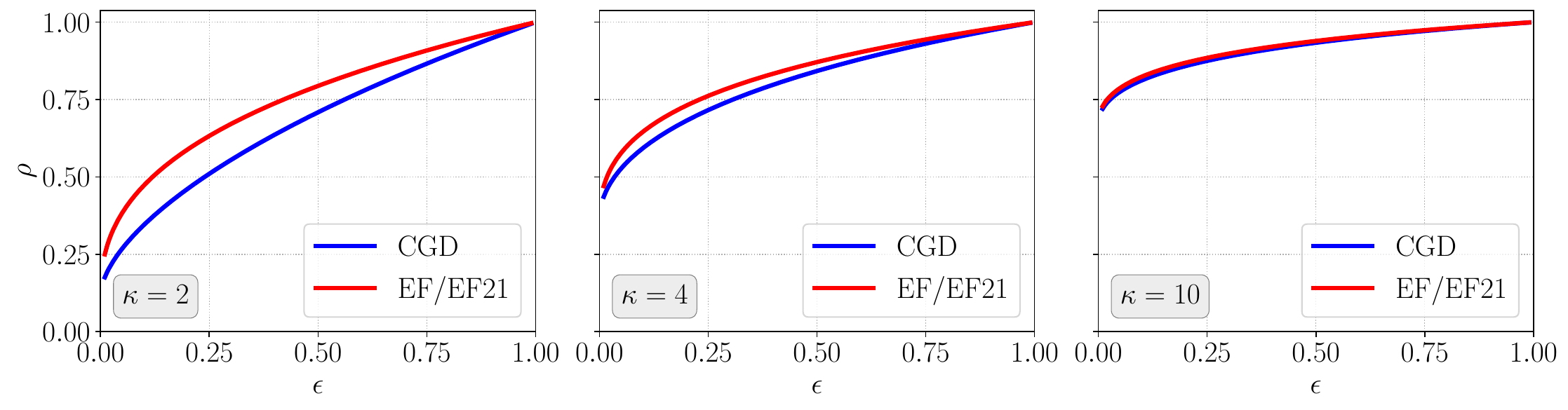}
    \caption{Line plot comparing the convergence rate of this $\CGD$ (blue) with $\EF$/$\EFtw$ (red) for various $\kappa$.}
    \label{fig:rate_comparison_cgd}
\end{figure}

In this section, we compare the rate given in~\cref{eq:rate_explicit} with the convergence rate of $\CGD$ and the standard convergence rate of $\EFtw$ \cite[see Theorem 2]{richtarik_ef21_2021}. We show that $\CGD$ is strictly superior to $\EF$/$\EFtw$ using a CAS certificate, that our rate is strictly better than the existing rate for $\EFtw$, and then we visualize the gap in both cases in~\Cref{fig:rate_comparison_rictharik,fig:rate_comparison_cgd}.

\begin{figure}[h]
    \centering
    \includegraphics[width=\textwidth]{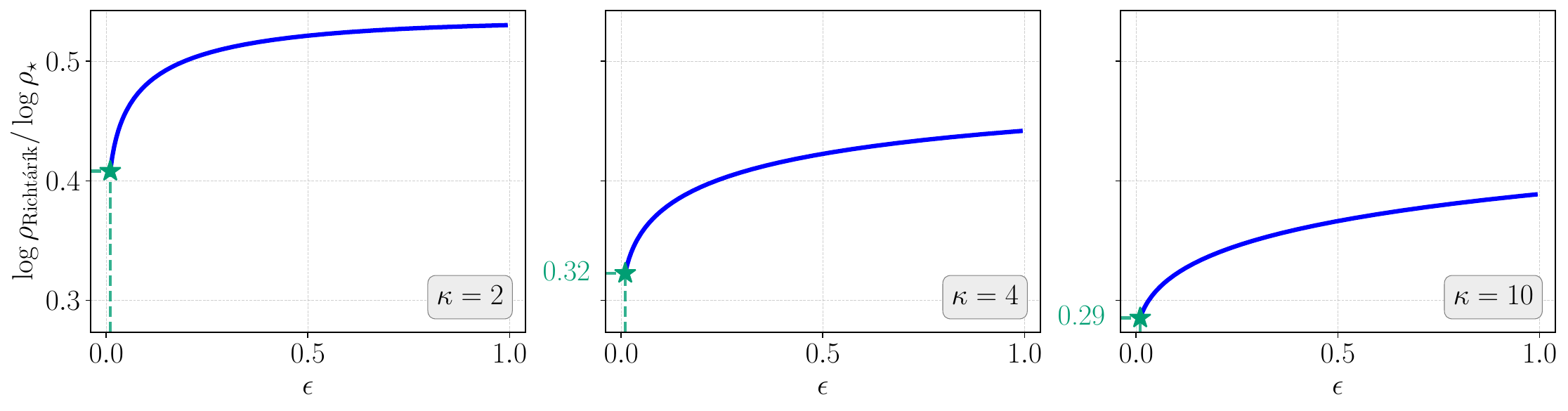}
    \caption{Line plot comparing the complexity of~\cref{eq:rho_richtarik} with the rates of this paper for various $\kappa$.}
    \label{fig:complexity_richtarik}
\end{figure}

\begin{figure}[h]
    \centering
    \includegraphics[width=\textwidth]{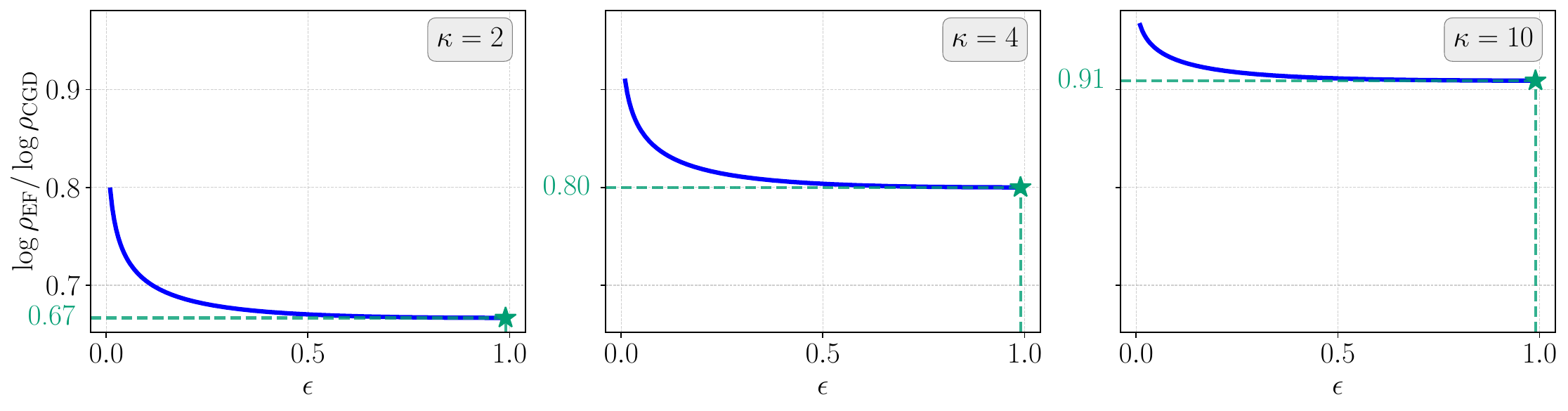}
    \caption{Line plot comparing the complexity of $\EF$/$\EFtw$ with $\CGD$ for various $\kappa$.}
    \label{fig:complexity_cgd_vs_ef}
\end{figure}

We also compare the "complexity" resulting from either method in \Cref{fig:complexity_richtarik,fig:complexity_cgd_vs_ef}. In order to create a fair and interpretable comparison we used the following metric:
\begin{equation}\label{eq:log_ratio_complexity}
\frac{\log \rho_A}{\log \rho_B}.
\end{equation}
The log-ratio can be motivated by the following example: if $\rho_A = 0.99$ and $\rho_B = 0.98$, then the latter will require half the number of iterations of the former before $\Lyap{\xi_N}{x_N}{f} \leq \epsilon$. 

\subsection{Comparison between $\CGD$ and $\EF$/$\EFtw$}\label{sec:comp_cgd_ef}
We now compare the convergence rate when using the optimal step size between $\CGD$ and $\EF$ (equivalently, $\EFtw$ since the rate is the same). Note that the optimal rate for $\CGD$ is given by \cite{gannot2022frequency, de2017worst}
\begin{equation}
\rho_{\CGD} = \left(\frac{\kappa_{\sqrt{\epsilon}} - 1}{\kappa_{\sqrt{\epsilon}} + 1} \right)^2,
\end{equation}
where $\kappa_{\sqrt{\epsilon}} = \kappa \left(\frac{1 + \sqrt \epsilon}{1 - \sqrt \epsilon}\right)$, when using step size \cite{gannot2022frequency}
\begin{equation}
\eta_{\CGD} = \frac{2}{(1 - \sqrt{\epsilon})\mu + (1 + \sqrt{\epsilon}) L}.
\end{equation}
The next statement has been verified using a WolframScript:
\certifiedbox[cas={https://github.com/DanielBergThomsen/error-feedback-tight/blob/main/certificates/CGD_superiority.wls}]{
\begin{align*}
\rho_{\EF} - \rho_{\CGD} =  \sqrt{\epsilon} +& \left(\frac{1 - \sqrt{\epsilon}}{2}\right) \left(\frac{\kappa - 1}{\kappa + 1}\right)^2 \left[1 - \sqrt{\epsilon} + \sqrt{(1 + \sqrt{\epsilon})^2 + \sqrt{\epsilon} 16 \frac{\kappa}{(\kappa - 1)^2}}\right] \\
-& \left(\frac{\kappa_{\sqrt{\epsilon}} - 1}{\kappa_{\sqrt{\epsilon}} + 1} \right)^2 > 0.
\end{align*}}

\subsection{Comparison to \citet{richtarik_ef21_2021}}\label{sec:analytical_comp_richtarik}
The rate of~\cite{richtarik_ef21_2021} for $\EFtw$ under the Łojasiewicz inequality, and with the largest possible step size chosen, is given by
\begin{equation}\label{eq:rho_richtarik}
\rho' \coloneqq \max\left\{1 - \frac{1 - \sqrt{\epsilon}}{2}, \ 1 - \kappa^{-1} \left(\frac{1 - \sqrt{\epsilon}}{1 + \sqrt{\epsilon}}\right) \right\}
\end{equation}
in the single-agent case. We begin by noting that to show that our rate is faster, we only have to show that it is smaller than one of the terms of~\cref{eq:rho_richtarik}.

\certifiedbox[cas={https://github.com/DanielBergThomsen/error-feedback-tight/blob/main/certificates/comparison_richtarik.wls}]{
Note that 
\begin{align*}
\rho' - \rho &=
1 - \sqrt{\epsilon}
- \frac{1 - \sqrt{\epsilon}}{(1+\sqrt{\epsilon})\,\kappa} \\
&- \frac{(1-\sqrt{\epsilon})(\kappa-1)^2\left(1 - \sqrt{\epsilon} + \sqrt{\,1 + \epsilon + \dfrac{2\sqrt{\epsilon}\,\bigl(1 + \kappa(6+\kappa)\bigr)}{(\kappa-1)^2}}\,\right)}{2(\kappa+1)^2}
\end{align*}
Writing this as a single fraction, we see that the sign of this expression is the same as that of
\begin{align*}
\Delta \coloneqq \bigl(1-\sqrt{\epsilon}\bigr)\Bigl(
  -2 + \kappa\bigl(
    -3 + \epsilon(\kappa-1)^2
    + 2\sqrt{\epsilon}(\kappa+1)^2
    + (1+\sqrt{\epsilon})R \\
    \qquad\qquad\quad
    + \kappa\bigl(4+\kappa - (1+\sqrt{\epsilon})R\bigr)
  \bigr)
\Bigr).
\end{align*}
where $R \coloneqq \sqrt{\,(\kappa-1)^2 + \epsilon(\kappa-1)^2 + 2\sqrt{\epsilon}\,\bigl(1+\kappa(6+\kappa)\bigr)}$. Denote the coefficient in front of $R$ as $C(R)$, and note that is negative. As a result, it just remains to make sure that the rest of the terms compensate for this term. Rewriting this as an inequality, and squaring to deal with $R$, our result follows from the inequality
$$
\left(\Delta - C(R) \cdot R\right)^2 - \left(C(R) \cdot R\right)^2 = 4 (1 - \sqrt{\epsilon})^2(1 + \kappa) F > 0,
$$
where
$$
F = 1 + \kappa\left(1 + \kappa(-5 + 3\kappa) + \epsilon(1 + \kappa)(-1 + 3\kappa) + 2\sqrt{\epsilon}\,(-1 + \kappa)(1 + 3\kappa)\right).
$$
The factors other than $F$ are clearly positive, and F is increasing with respect to $\epsilon$. Setting $\epsilon$ to $0$ shows that it is positive for all valid $\epsilon$.}

\clearpage

\section{Missing proofs}
\label{sec:proofs}
This section contains the proofs of \Cref{thm:ef,thm:ef21}.
The proofs were obtained using by observing numerical results using the Lyapunov search procedure presented in \Cref{a:EF_feas,a:EF21_feas}. The resulting proofs are remarkably compact, and the main technical step consists of verifying an algebraic reformulation by hand. For ease of verification, we provide the reader with scripts written in Wolfram Language\footnote{Wolfram Language is the programming language used in Mathematica. The scripts can be used to verify our rates without a paid license using Wolfram Engine.} that automatically performs those reformulations using a computer algebra system.

\subsection{Proof of Theorem~\ref{thm:ef}}
\theoremEF*

\label{sec:proof_ef}
\begin{proof}
    We begin by proving the rate given in~\eqref{eq:ef_recurrence} for our Lyapunov function. 
      Consider the following inequalities, and associated with each of them the assigned multiplier 
      \footnote{These multipliers correspond to closed forms for some of the variables of \eqref{eq:ef_sdp} when the Lyapunov function is fixed as in the statement of the theorem.}
    \begin{equation*}
        \begin{aligned}
            \begin{aligned}
            I_{\F_{\mu, L}}^{(1)} \coloneqq f(x_k) - \fs - \Df(x_k)^\top (x_k - \xs) + \frac{1}{2L} \|\Df(x_k)\|^2 \\
            + \frac{\mu}{2(1 - \mu / L)} \|x_k - \xs - \frac{1}{L} \Df(x_k)\|^2 \leq 0,
            \end{aligned}
            & \quad & : \lambda \\
            I_{\F_{\mu, L}}^{(2)} \coloneqq \fs - f(x_k) + \frac{1}{2L} \|\Df(x_k)\|^2 + \frac{\mu}{2(1 - \mu / L)} \|x_k - \xs - \frac{1}{L} \Df(x_k)\|^2 \leq 0,
            & \quad & : \lambda \\
            I_{\C} \coloneqq \|e_{k+1}\|^2 - \epsilon \|e_k + \eta \Df(x_k)\|^2 \leq 0, & \quad & : \nu \\
        \end{aligned}
    \end{equation*}
    where \(\lambda\) is defined in \eqref{eq:lagrange_ef}, and \(\nu \coloneqq \frac{1}{\sqrt{\epsilon}}\). 

    \certifiedbox[cas={https://github.com/DanielBergThomsen/error-feedback-tight/blob/main/certificates/EF/upper_bound_CAS.wls}]{
        Summing these inequalities with their multipliers, plugging in the update rules for 
    \(x_{k+1}\) and \(e_{k+1}\), and using \(\rho\) to denote the contraction factor we got in 
    \eqref{eq:ef_recurrence}, we can rewrite the resulting inequality as: 
    \begin{equation}
        \rho \cdot \Vbase(x_k, e_k) \geq \Vbase(x_{k+1}, e_{k+1}) + a \cdot \|e_k - 
        \frac{\rho - 1}{a} (x_k - \xs) + \frac{2(\sqrt{\epsilon} - 1)}{a(L + \mu)} g_k\|^2,
    \end{equation}
    where 
    \begin{equation}
        a \coloneqq (\rho - \sqrt{\epsilon}) \cdot \left(\frac{1 + \sqrt{\epsilon}}{\sqrt{\epsilon}}\right).
    \end{equation}}
    The statement now follows from the simple inequality \(\rho > \sqrt{\epsilon}\).

    We now prove that the announced rate is tight. Consider the one-dimensional quadratic 
    function
    \begin{equation}
        f_\mu(x) = \frac{\mu}{2} x^2.
    \end{equation}
    The proof strategy used here is to show that the contraction for our Lyapunov
    function asymptotically matches the convergence rate announced in Theorem~\ref{thm:ef}.
    We begin by fully exploiting Assumption~\ref{as:compression} and set
    \begin{equation}
        c_k \coloneqq \C(e_k + \eta \Df(x_k)) = (1 + \sqrt{\epsilon}) \cdot (\eta \Df(x_k) + e_k).
    \end{equation}
    We can now rewrite the update rule for \(x_{k+1}\) and \(x_{k+2}\) to get an expression 
    for \(e_k\) and \(e_{k+1}\) respectively:
    \begin{equation}
        e_k = \frac{1 - \eta \mu (1 + \sqrt{\epsilon})}{1 + \sqrt{\epsilon}} x_k - \frac{x_{k+1}}{1 + \sqrt{\epsilon}}, \quad
        e_{k+1} = \frac{1 - \eta \mu (1 + \sqrt{\epsilon})}{1 + \sqrt{\epsilon}} x_{k+1} - \frac{x_{k+2}}{1 + \sqrt{\epsilon}},
    \end{equation}
    after which we use the update rule for \(e_{k+1}\) of Algorithm~\ref{alg:ef} to get a second-order recurrence 
    relation for the sequence \(\{x_k\}_{k=1}^{\infty}\):
    \begin{equation}
        \sqrt{\epsilon} x_k = x_{k+2} - (1 - \eta \mu - \sqrt{\epsilon} (1 + \eta \mu)) \cdot x_{k+1}
    \end{equation}
    The solution to this recurrence relation is given by the roots of the characteristic equation, 
    and after plugging in the initial conditions, we get
    \begin{equation}\label{eq:lb_asymptotic_x}
        \begin{aligned}
            x_k & = \frac{1}{T} \cdot (1 - \eta \mu + \sqrt{\epsilon} (1 - \eta \mu) + T)
            (1 - \eta \mu - \sqrt{\epsilon} (1 + \eta \mu) + T)^k \\
            & - \frac{1}{T} \cdot (1 - \eta \mu + \sqrt{\epsilon} (1 - \eta \mu) - T)
            (1 - \eta \mu - \sqrt{\epsilon} (1 + \eta \mu) - T)^k,
        \end{aligned}
    \end{equation}
    where \(T \coloneqq \sqrt{4 \sqrt{\epsilon} + (1 - \eta \mu + \sqrt{\epsilon} (1 + \eta \mu))^2}\). 
    Note that for 
    \begin{equation}
        \eta < \left( \frac{1}{\mu} \right) \cdot \left(\frac{1 - \sqrt{\epsilon}}{1 + \sqrt{\epsilon}}\right),
    \end{equation}
    which is strictly larger than the step size given in \eqref{eq:ef_step_size}, 
    the above expression is dominated by the first term in the limit \(k \to \infty\).
    If we plug in the resulting asymptotic expression for \(x_k\) into the definition of 
    \(e_k\), and plug the resulting points into our Lyapunov function we get
    \begin{equation} \label{eq:lb_asymptotic_V}
        \frac{\Vbase(x_{k+1}, e_{k+1})}{\Vbase(x_k, e_k)} \xrightarrow{k \to \infty} \frac{1}{4} (1 - \eta \mu - 
        \sqrt{\epsilon} (1 + \eta \mu) + T)^2
    \end{equation}
    which, after plugging in the step size given in \eqref{eq:ef_step_size}, is exactly the convergence rate 
    announced in Theorem~\ref{thm:ef}. The fact that our Lyapunov function is tight now follows 
    from the remark made in Section~\ref{sec:tightness}.

    Finally, we prove that the step size given in \eqref{eq:ef_step_size} is the optimal 
    step size for our Lyapunov function. Note that the contraction factor
    \begin{equation}
        \rho(\eta) \coloneqq \frac{1}{4} (1 - \eta \mu - \sqrt{\epsilon} (1 + \eta \mu) + T)^2,
    \end{equation}
    that becomes the dominant term in the limit \(k \to \infty\) of~\eqref{eq:lb_asymptotic_V}
    is strictly decreasing in \(\eta\). This is immediate from inspecting the sign of the
    derivative of the expression with respect to \(\eta\):
    \begin{equation}
        \frac{d \rho(\eta)}{d \eta} = -\mu(1+\sqrt{\epsilon})\frac{(1 - \eta \mu - \sqrt{\epsilon} 
        (1 + \eta \mu) + T)^2}{2 T}.
    \end{equation}
    The rest of the proof now follows from instead considering the quadratic given by
    \begin{equation}
        f_L(x) \coloneqq \frac{L}{2} x^2,
    \end{equation}
    and repeating all the arguments stated above, except we instead consider step sizes
    \begin{equation}
        \eta > \left(\frac{1}{L} \right) \cdot \left(\frac{1 - \sqrt{\epsilon}}{1 + \sqrt{\epsilon}}\right),
    \end{equation}
    and show that the contraction for our Lyapunov function is strictly decreasing for these
    step sizes. The argument now follows from the fact that the contraction factor on both
    of these quadratics are the same for the step size given in \eqref{eq:ef_step_size}.
\end{proof}

\subsection{Proof of Theorem~\ref{thm:ef21}}
\theoremEFtw*

\label{sec:proof_ef21}
\begin{proof}
    We begin by proving the rate given in~\eqref{eq:ef21_recurrence} for our Lyapunov function. 
    Consider the following inequalities, and associated with each of them the assigned multiplier:
    \begin{equation*}
        \begin{aligned}
            \begin{aligned}
            I_{\F_{\mu, L}}^{(1)} \coloneqq f(x_k) - f(x_{k+1}) + \frac{\|\Df(x_{k+1}) - \Df(x_k)\|^2}{2L} + \Df(x_k)^\top (x_{x+1} - x_k) \\
            + \frac{\mu}{2(1 - \mu / L)} \|x_k - x_{k+1} - \frac{1}{L} (\Df(x_k) - \Df(x_{k+1}))\|^2 \leq 0,
            \end{aligned}
            & \quad & : \lambda' \\
            \begin{aligned}
            I_{\F_{\mu, L}}^{(2)} \coloneqq f(x_{k+1}) - f(x_k) + \frac{\|\Df(x_k) - \Df(x_{k+1})\|^2}{2L} + \Df(x_{k+1})^\top (x_k - x_{k+1}) \\
            + \frac{\mu}{2(1 - \mu / L)} \|x_{k+1} - x_k - \frac{1}{L} (\Df(x_{k+1}) - \Df(x_k))\|^2 \leq 0,
            \end{aligned}
            & \quad & : \lambda' \\
            I_{\C} \coloneqq \|\Df(x_{k+1}) - d_k - \C(\Df(x_{k+1}) - d_k)\|^2 - \epsilon\|\Df(x_{k+1}) - d_k\|^2 \leq 0, & \quad & : \nu \\
        \end{aligned}
    \end{equation*}
    where \(\lambda'\) is defined as
    \begin{equation}\label{eq:lambda_ef21}
    \lambda' \coloneqq \frac{\sqrt{\epsilon}}{\eta^\star (L + \mu)} \left[ (1 - \sqrt{\epsilon})(L - \mu) + (1 + \sqrt{\epsilon}) \sqrt{(L - \mu)^2 + \frac{16 L \mu \sqrt{\epsilon}}{(1 + \sqrt{\epsilon})^2}} \right],
    \end{equation}
    and \(\nu \coloneqq 1\).

    \certifiedbox[cas={https://github.com/DanielBergThomsen/error-feedback-tight/blob/main/certificates/EF21/upper_bound_CAS.wls}]{
    Summing these inequalities with their multipliers, plugging in the update rules for 
    \(x_{k+1}\) and \(d_{k+1}\), and using \(\rho\) to denote the contraction factor we got in 
    \eqref{eq:ef21_recurrence}, we can rewrite the resulting inequality as:
    \begin{equation}
        \rho \cdot \Vbase(g_k, d_k) \geq \Vbase(g_{k+1}, d_{k+1}) + a \cdot \|d_k + 
        \frac{1}{a} ((\epsilon + b) g_{k+1} - (\rho + b) g_k)\|^2,
    \end{equation}
    where 
    \begin{equation}
        b \coloneqq \frac{\lambda}{L - \mu} \cdot \left(\frac{1 - \sqrt{\epsilon}}{1 + \sqrt{\epsilon}}\right).
    \end{equation}
    and
    \begin{equation}
        a \coloneqq \rho - \epsilon + 2\eta \lambda (1 - \sqrt{\epsilon}) \frac{L \mu}{L - \mu}.
    \end{equation}}
    The statement now follows from plugging in the value of our multipliers and checking the sign.

    We now prove that the announced rate is tight. Consider the one-dimensional quadratic 
    function
    \begin{equation}
        f_\mu(x) = \frac{\mu}{2} x^2.
    \end{equation}
    The proof strategy used here is to show that the contraction for our Lyapunov
    function asymptotically matches the convergence rate announced in Theorem~\ref{thm:ef21}.
    We begin by fully exploiting Assumption~\ref{as:compression} and set
    \begin{equation}
        c_k \coloneqq \C(\Df(x_{k+1}) - d_k) = (1 + \sqrt{\epsilon}) \cdot (\Df(x_{k+1}) - d_k)
    \end{equation}
    We can now rewrite the update rule for \(x_{k+1}\) and \(x_{k+2}\) to get an expression 
    for \(d_k\) and \(d_{k+1}\) respectively:
    \begin{equation}
        d_k = \frac{x_k - x_{k+1}}{\eta}, \quad
        d_{k+1} = \frac{x_{k+1} - x_{k+2}}{\eta},
    \end{equation}
    after which we use the update rule for \(d_{k+1}\) of Algorithm~\ref{alg:ef21} to get a second-order recurrence 
    relation for the sequence \(\{x_k\}_{k=1}^{\infty}\):
    \begin{equation}
        \sqrt{\epsilon} x_k = x_{k+2} - (1 - \eta \mu - \sqrt{\epsilon} (1 + \eta \mu)) \cdot x_{k+1}. 
    \end{equation}
    Note that this is the exact same recurrence relation as in \eqref{eq:ef_recurrence}, which
    means we can reuse the expression in~\eqref{eq:lb_asymptotic_x} and the argument that follows.
    The proof now follows from the definition of our Lyapunov function in~\eqref{eq:ef21_lyapunov}.
    The optimality of our Lyapunov function similarly follows from the same argument given in the remark
    made in Section~\ref{sec:methodology}.

    Lastly, the proof of optimality of our step size follows directly from the same argument
    given in the proof of Theorem~\ref{thm:ef}.
\end{proof}

\subsection{Extension to stochastic compressors} \label{sec:stoch_extension}
Note that all the results of \Cref{thm:ef,thm:ef21} were proven using the deterministic version of \Cref{as:compression}. Any deterministic compressor satisfies that assumption, and our lower bounds thus remain valid in this case. To show that the given rates are tight, we have to replace the inequalities $I_\C$ from the respective proofs with their stochastic counterparts:
\begin{align*}
    I_{\C}^{\EF} \coloneqq \mathbb{E}\left[\|e_{k+1}\|^2\right] - \epsilon \|e_k + \eta \Df(x_k)\|^2 \leq 0, \\
    I_{\C}^{\EFtw} \coloneqq \mathbb{E}\left[\|\Df(x_{k+1}) - d_k - \C(\Df(x_{k+1}) - d_k)\|^2\right] - \epsilon\|\Df(x_{k+1}) - d_k\|^2 \leq 0.
\end{align*}
\certifiedbox[cas=https://github.com/DanielBergThomsen/error-feedback-tight/blob/main/certificates/EF/upper_bound_random_CAS.wls]{For $\EF$, we need to show the following:
\begin{align*}
\rho \cdot \mathbb{E} \Vbase(x_k, e_k) \geq \mathbb{E} \Vbase(x_{k+1}, e_{k+1}),
\end{align*}}

\certifiedbox[cas=https://github.com/DanielBergThomsen/error-feedback-tight/blob/main/certificates/EF21/upper_bound_random_CAS.wls]{And for $\EFtw$ we need to show the following:
\begin{align*}
\rho \cdot \mathbb{E} \Vbase(g_k, d_k) \geq \mathbb{E} \Vbase(g_{k+1}, d_{k+1}).
\end{align*}}
The same exact proof as used in the deterministic case holds for both due to the linearity of expectation. Wolfram Language scripts verifying this fact are available in the source code repository for this article.


\newpage
\section{Additional numerical results} \label{app:additional_numerical_results}
All subsections include details on how the corresponding results were computed. This section is organized as 
follows:
\begin{itemize}
    \item \Cref{sec:performance_plots} presents additional performance plots for 
    all methods and explains how the plots in the main paper were generated.
    \item \Cref{sec:lyapunov_multistep} provides illustrations demonstrating that our Lyapunov 
    functions remain tight over multiple iterations.
    \item \Cref{sec:lyapunov_tightness} includes additional tables further confirming the tightness of our 
    Lyapunov functions.
    \item \Cref{sec:step_size} shows plots illustrating the optimality of our step size.
    \item \Cref{sec:lyapunov_compare} numerically demonstrates that the type of Lyapunov function used in \citet{richtarik_ef21_2021} achieves a worse rate of convergence than the one used in this work.
\end{itemize}
All experiments were run on a MacBook Pro with an M4 Max processor. While none of the experiments are computationally demanding by modern standards, they can be scaled by increasing the resolution of the $\eta$ and $\epsilon$ grid to produce finer plots.

The source code for all the experiments is publically available in the following GitHub repository: \href{https://github.com/DanielBergThomsen/error-feedback-tight}{https://github.com/DanielBergThomsen/error-feedback-tight} 

\subsection{Performance plots}
\label{sec:performance_plots}
This section presents the worst-case performance of all methods studied in this work,
plotted as a function of the step size \(\eta\) and the compression parameter \(\epsilon\).

\begin{figure}[h]
    \centering
    \includegraphics[width=\textwidth]{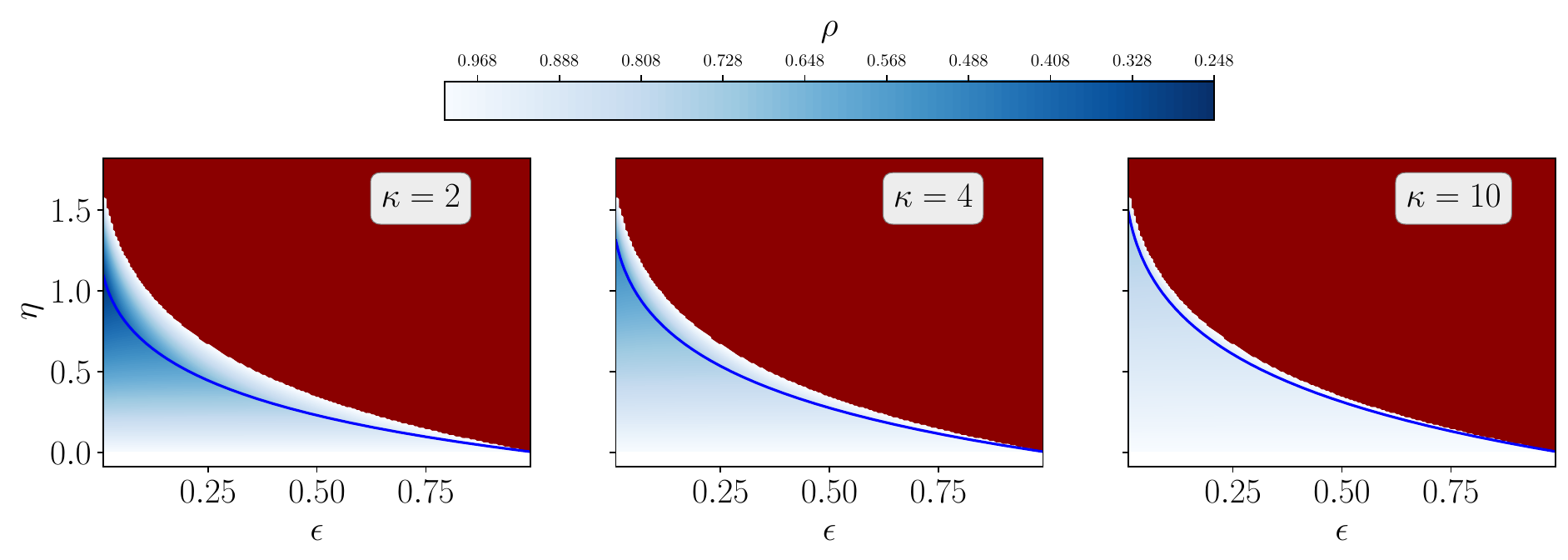}
    \caption{Contour plot showing the performance of $\EF$ as a function of step size $\eta$ and compression 
    parameter $\epsilon$, with regions of non-convergence marked in red. The regions of non-convergence were 
    identified using PEPit by finding cycles of length 2. Each column corresponds to $\mu = 0.5, 0.25, 0.1$,
    with $L=1.0$ fixed across all plots. The optimal step size setting for a given \(\epsilon\) is marked in blue.}
    \label{fig:ef_performance}
\end{figure}

All contour plots were evaluated over a grid with \(\epsilon \in [0.01, 0.99]\), and 
\(\eta \in [0.01, \frac{2}{L + \mu_\star}]\), where \(\mu_\star\) is the smallest \(\mu\) specified 
in the caption of each figure (except for Figure~\ref{fig:ef_performance}, where it is set to \(0.1\)).
Each axis was discretized with a resolution of 200 points.

To generate each non-cyclic point, we used the following procedure:
\begin{enumerate}
    \item For each method, we computed the optimal Lyapunov function (without additional constraints)
    via bisection on the contraction factor \(\rho\), up to a precision of \(10^{-6}\).
    \item Using the resulting Lyapunov function, we then computed the worst-case contraction factor using 
    PEPit~\cite{goujaud2022pepit} and the MOSEK solver~\cite{mosek}.
\end{enumerate}
We adopt this two-step approach because the feasibility problems used to 
compute the Lyapunov functions suffer from numerical instability. By evaluating the contraction factor 
separately using PEPit, we ensure that the reported value is an upper bound on the true contraction 
factor---up to solver tolerance.

To identify the area of non-convergence in the plots, we check whether a cycle exists for each
pair of \(\eta\) and \(\epsilon\). This is done by following the procedure
outlined in \citet{goujaud2023counter}: we compute the worst-case 
performance of the metric \(- \|x_k - x_0\|^2\) for CGD, \(-\|x_k - x_0\|^2 - \|e_k - e_0\|^2\) for \(\EF\), and 
\(-\|x_k - x_0\|^2 - \|d_k - d_0\|^2\) for \(\EFtw\). If this value falls below a threshold
(set to \(10^{-3}\)), we conclude that a cycle is present. In our experiments,
cycles of length 2 were successfully identified for all methods, and these matched precisely with the regions
of the contour plots where \(\rho > 1\).

\begin{figure}[h]
    \centering
    \includegraphics[width=\textwidth]{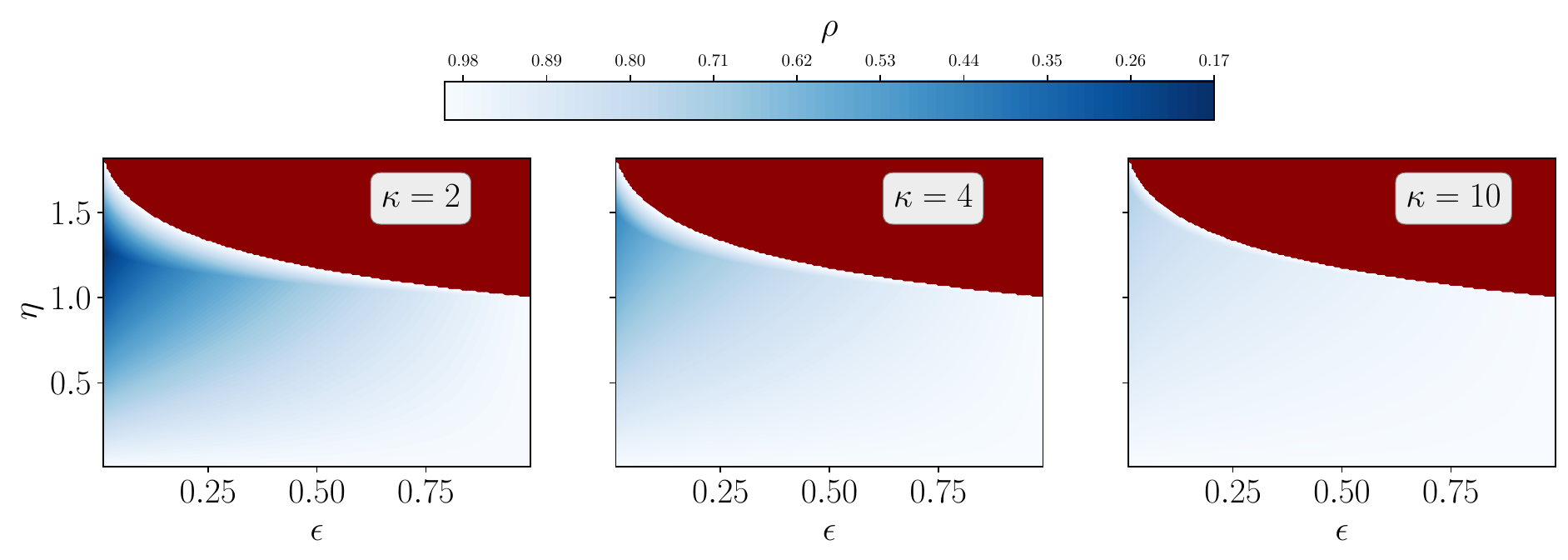}
    \caption{Contour plot showing the performance of $\CGD$ as a function of step size $\eta$ and compression 
    parameter $\epsilon$, with regions of non-convergence marked in red. The regions of non-convergence were 
    identified using PEPit by finding cycles of length 2. Each column corresponds to $\mu = 0.5, 0.25, 0.1$,
    with $L=1.0$ fixed across all plots.}
    \label{fig:cgd_performance}
\end{figure}

\begin{figure}[h]
    \centering
    \includegraphics[width=\textwidth]{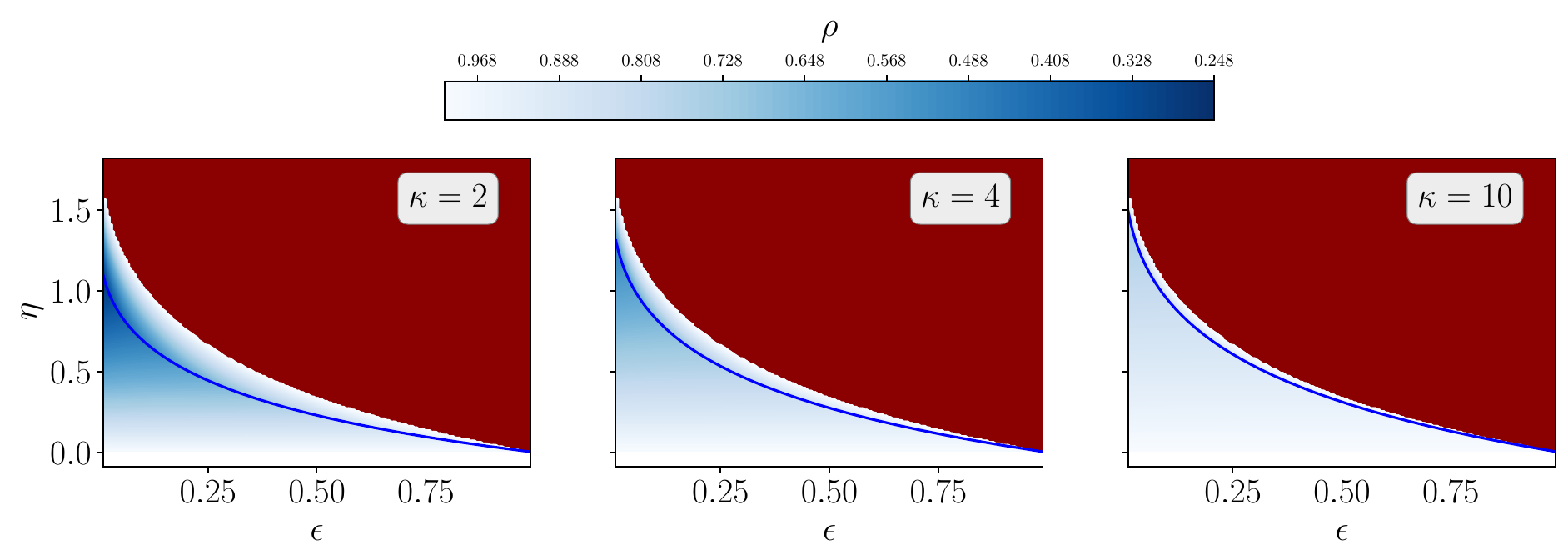}
    \caption{Contour plot showing the performance of $\EFtw$ as a function of step size $\eta$ and compression 
    parameter $\epsilon$, with regions of non-convergence marked in red. The regions of non-convergence were 
    identified using PEPit by finding cycles of length 2. Each column corresponds to $\mu = 0.5, 0.25, 0.1$,
    with $L=1.0$ fixed across all plots. The optimal step size setting for a given \(\epsilon\) is marked in blue.}
    \label{fig:ef21_performance}
\end{figure}

\subsection{Multi-step Lyapunov analysis}
\label{sec:lyapunov_multistep}
\begin{figure}[h]
    \centering
    \begin{tabular}{ccc}
        \includegraphics[width=0.3\textwidth]{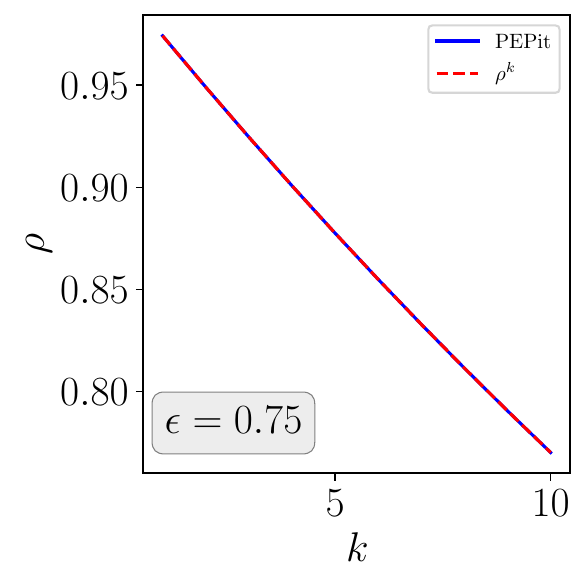} &
        \includegraphics[width=0.3\textwidth]{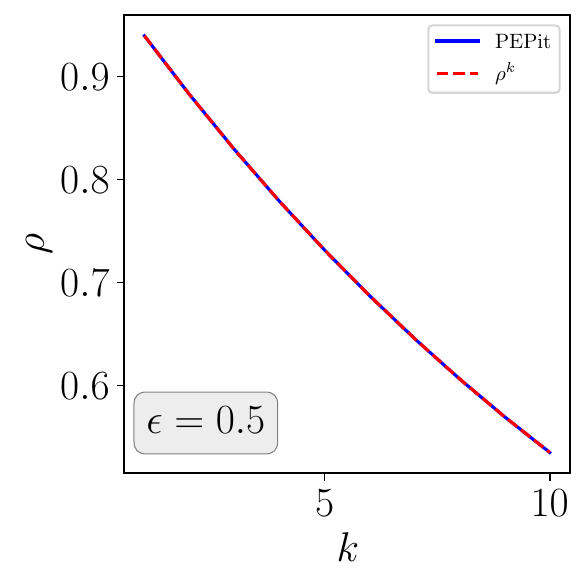} &
        \includegraphics[width=0.3\textwidth]{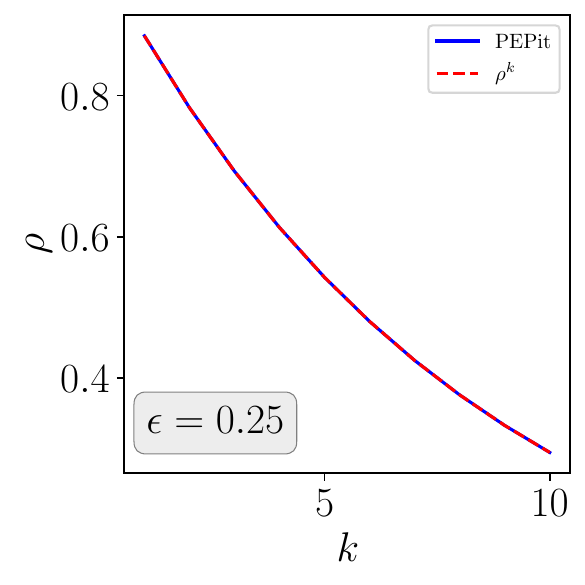}
    \end{tabular}
    \caption{Multi-step Lyapunov analysis for $\EF$, computed using PEPit. The blue line shows the 
    contraction factor achieved by the Lyapunov function after \(k\) iterations, while the red dashed line represents 
    the theoretical rate \(\rho^k\), where \(\rho\) is the single-step contraction factor. Each column 
    corresponds to a different value of \(\epsilon = 0.75, 0.5, 0.25\), with \(L=1.0\) and \(\mu=0.1\) fixed 
    across all plots.}
    \label{fig:lyapunov_multistep_ef}
\end{figure}

\begin{figure}[h]
    \centering
    \begin{tabular}{ccc}
        \includegraphics[width=0.3\textwidth]{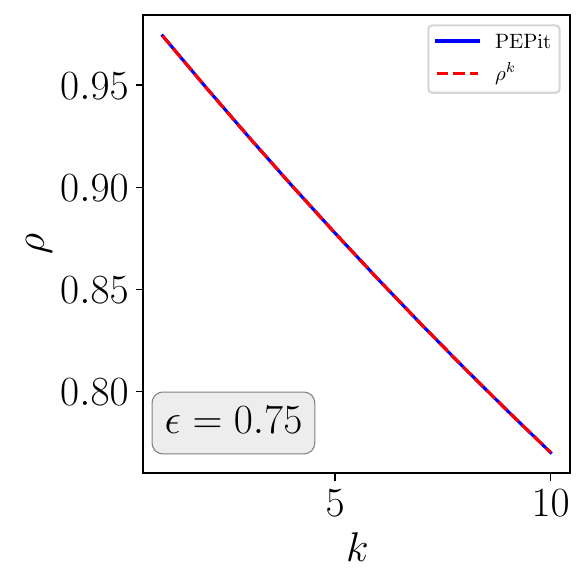} &
        \includegraphics[width=0.3\textwidth]{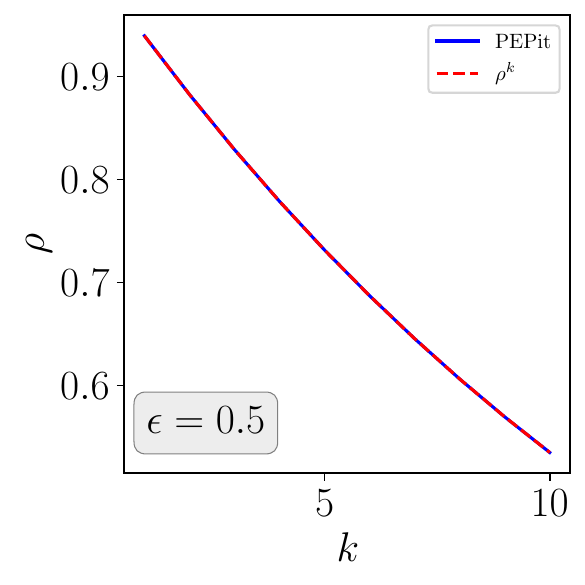} &
        \includegraphics[width=0.3\textwidth]{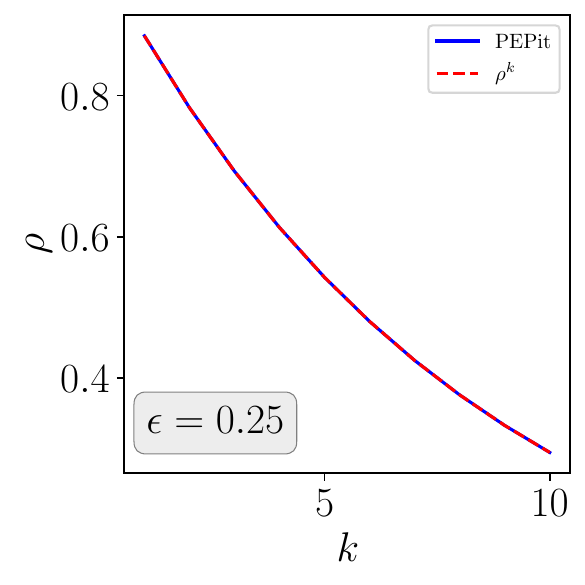}
    \end{tabular}
    \caption{Multi-step Lyapunov analysis for $\EFtw$, computed using PEPit. The blue line shows the 
    contraction factor achieved by the Lyapunov function after \(k\) iterations, while the red dashed line represents 
    the theoretical rate \(\rho^k\), where \(\rho\) is the single-step contraction factor. Each column 
    corresponds to a different value of \(\epsilon = 0.75, 0.5, 0.25\), with \(L=1.0\) and \(\mu=0.1\) fixed 
    across all plots.}
    \label{fig:lyapunov_multistep_ef21}
\end{figure}

In this section, we show that our simple Lyapunov functions achieve the claimed convergence rate
over multiple iterations. Specifically, we use PEPit to compute the contraction factor achieved by the Lyapunov
function after \(k\) iterations and compare it to the theoretical rate \(\rho^k\), where \(\rho\) is the 
single-step contraction factor. 
The exact match between these quantities in \Cref{fig:lyapunov_multistep_ef,fig:lyapunov_multistep_ef21} confirms that our single-step analysis accurately characterizes
the worst-case performance over multiple iterations on these Lyapunov functions.

\subsection{Lyapunov function class tightness}
\label{sec:lyapunov_tightness}
In this section, we show that for various conditioning numbers \(\kappa\), our Lyapunov
functions for $\EF$ and $\EFtw$ are tight with respect to our class of Lyapunov functions, when
using optimal step sizes. We remark that our Lyapunov functions are actually tight for many step size
settings, but notably not step sizes which are larger than the optimal step size.

The tables report the maximum absolute difference in contraction factors between our Lyapunov function and the optimal one, over a range of \(\epsilon\) and \(\eta\) values specified in the captions. All contraction factors were computed using PEPit, 
and the procedure for the uncontrained Lyapunov functions is the one outlined in \Cref{sec:performance_plots}. Points where either Lyapunov function yields a contraction
factor greater than 1 were excluded from the computation of the maximum absolute difference.

\begin{table}[h]
    \centering
    \begin{tabular}{lrrr}
\toprule
 & $\kappa = 2$ & $\kappa = 4$ & $\kappa = 10$ \\
\midrule
Absolute error & 3.70e-07 & 4.83e-07 & 6.60e-07 \\
\bottomrule
\end{tabular}

    \caption{Maximum absolute difference in contraction factor for $\EF$ 
    when comparing the general Lyapunov function---constructed using any combination of state terms 
    specified in \Cref{sec:numerical}---to the simplified Lyapunov function defined in Theorem~\ref{thm:ef}. 
    The results are computed over a line with \(\epsilon \in [0.01, 0.99]\) and with \(\eta\) set to the optimal step size
    for \(L = 1\), and \(\mu = 0.5, 0.25, 0.1\).}
    \label{tab:lyapunov_tightness_ef}
\end{table}

\begin{table}[h]
    \centering
    \begin{tabular}{lrrr}
\toprule
 & $\kappa = 2$ & $\kappa = 4$ & $\kappa = 10$ \\
\midrule
Absolute error & 2.77e-07 & 1.97e-06 & 1.65e-06 \\
\bottomrule
\end{tabular}

    \caption{Maximum absolute difference in contraction factor for $\EFtw$ 
    when comparing the general Lyapunov function---constructed using any combination of state terms 
    specified in \Cref{sec:numerical}---to the simplified Lyapunov function defined in Theorem~\ref{thm:ef21}. 
    The results are computed over a line with \(\epsilon \in [0.01, 0.99]\) and with \(\eta\) set to the optimal step size
    for \(L = 1\), and \(\mu = 0.5, 0.25, 0.1\).}
    \label{tab:lyapunov_tightness_ef21}
\end{table}

\subsection{Step size comparison}
\label{sec:step_size}
\begin{figure}[h]
    \centering
    \begin{tabular}{ccc}
        \includegraphics[width=0.3\textwidth]{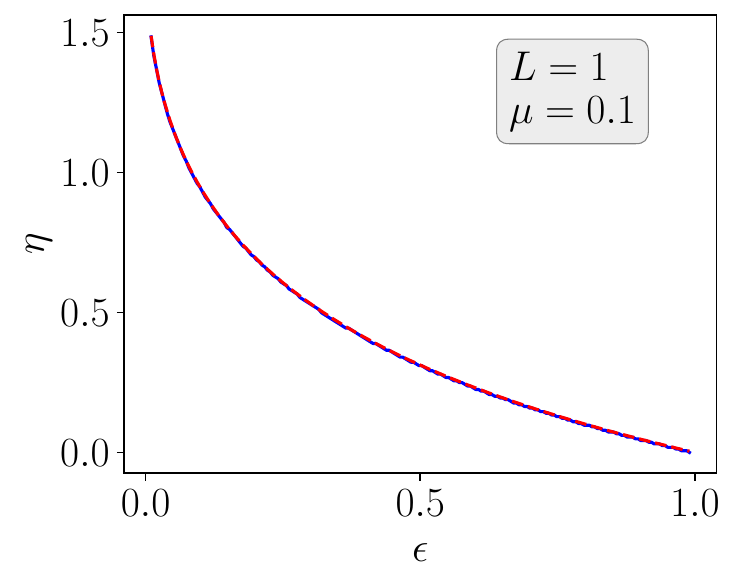} &
        \includegraphics[width=0.3\textwidth]{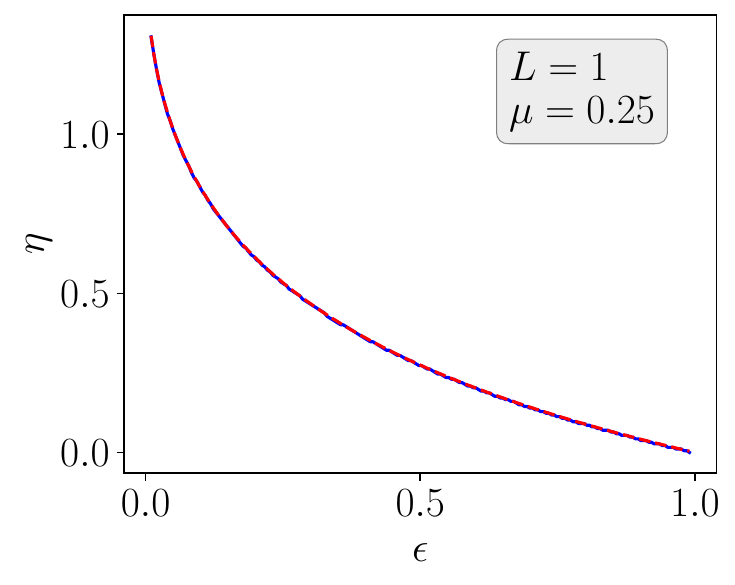} &
        \includegraphics[width=0.3\textwidth]{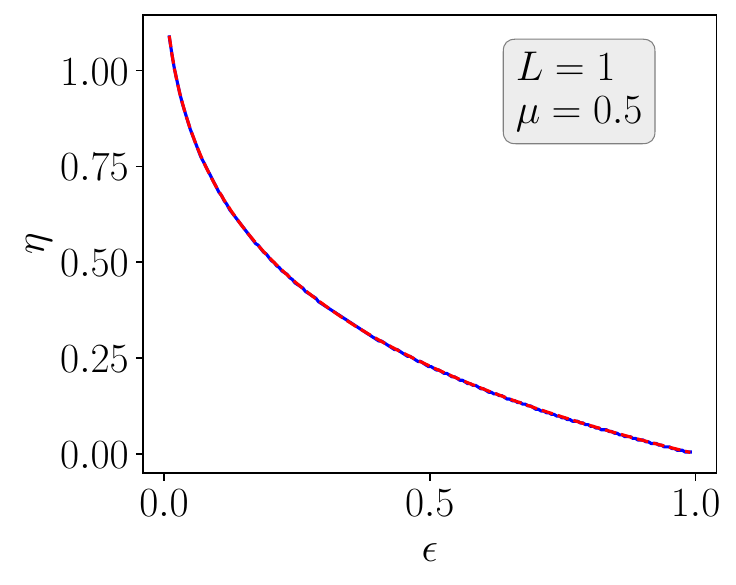} \\
        \includegraphics[width=0.3\textwidth]{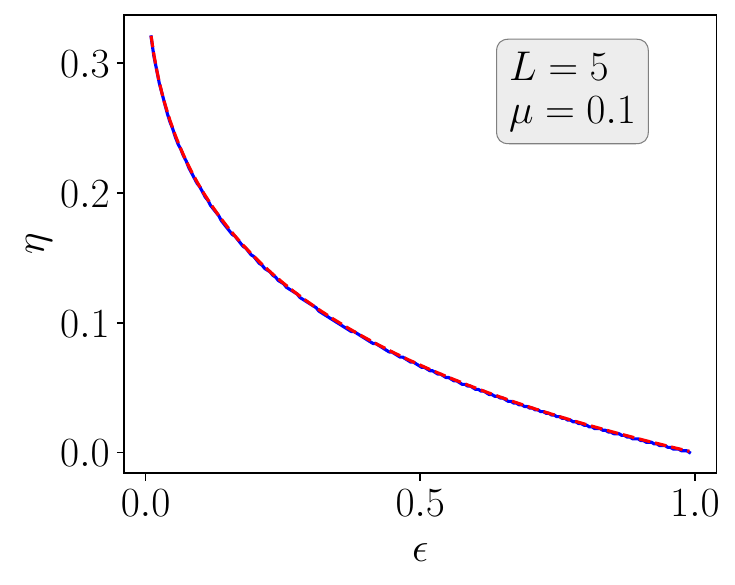} &
        \includegraphics[width=0.3\textwidth]{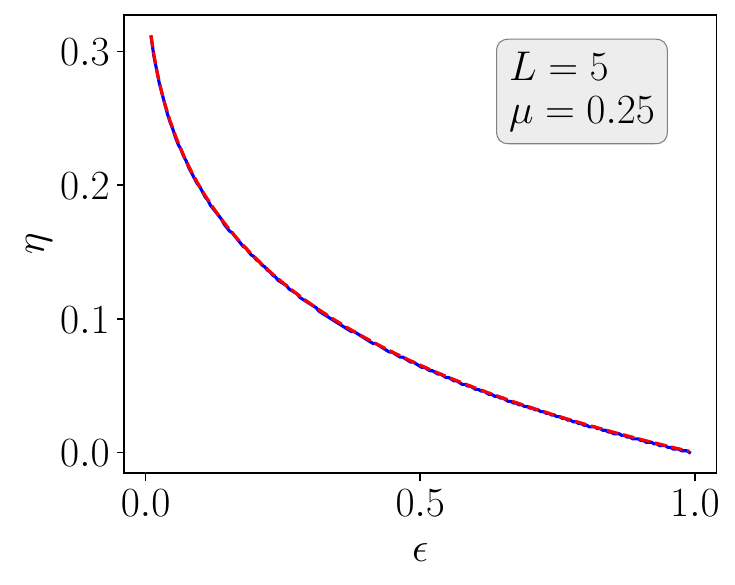} &
        \includegraphics[width=0.3\textwidth]{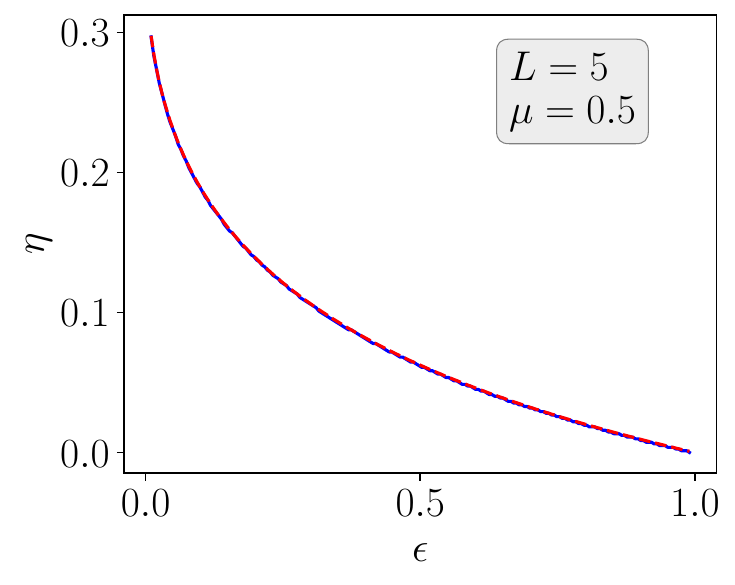} \\
        \includegraphics[width=0.3\textwidth]{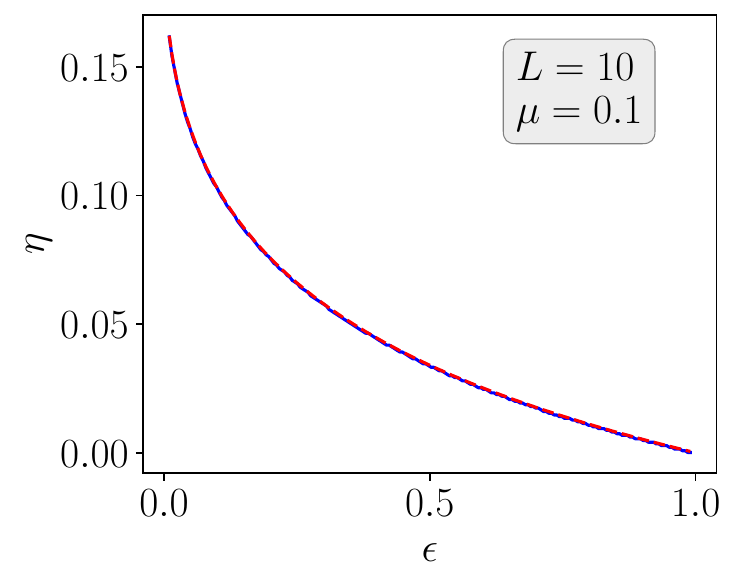} &
        \includegraphics[width=0.3\textwidth]{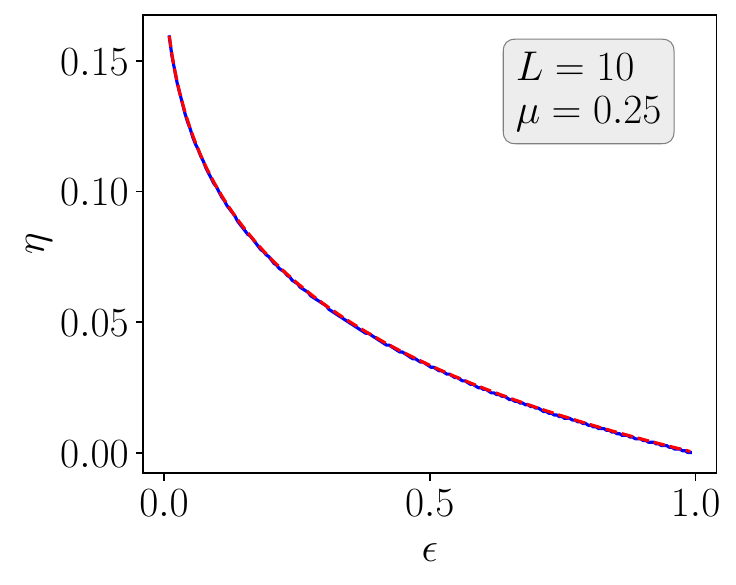} &
        \includegraphics[width=0.3\textwidth]{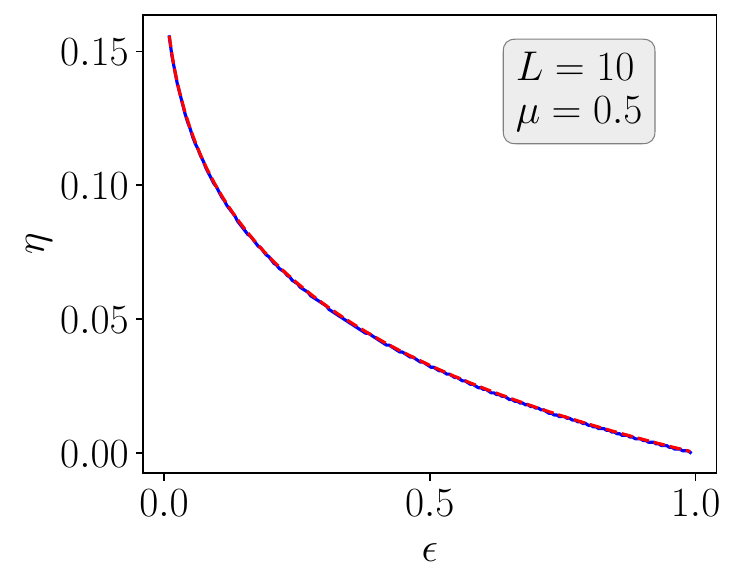} \\
    \end{tabular}
    \caption{Empirically observed optimal step sizes (blue) in comparison with 
    our setting (red, dashed) as a function of \(\epsilon\) for different values of \(\mu\) and \(L\).}
    \label{fig:ef_step_size}
\end{figure}

In this section, we compare the theoretically optimal step sizes we propose for our methods with empirically 
optimal step sizes determined through numerical experiments. To compute the empirical optima, we evaluate a grid of
\(\eta\) and \(\epsilon\) values and select the step size that minimizes the contraction factor achieved by our simplified 
Lyapunov functions. The results of that experiment are found in \Cref{fig:ef_step_size}.

\subsection{Comparison of Lyapunov functions to previous work}\label{sec:lyapunov_compare}
In this subsection, we compare the convergence rates achieved with our Lyapunov function compared to that of \citet{richtarik_ef21_2021}. For reference, the Lyapunov function they use is the following:
\begin{equation} \label{eq:lyapunov_richtarik}
    \Lyap{\xi^{\EFtw}}{x}{f} \coloneqq f(x) - f_\star + \frac{\eta}{\theta}\|d - g\|^2,
\end{equation}
where $\theta \coloneqq \frac{\epsilon}{1-\sqrt{\epsilon}}$. We ran an experiment where we measured the best possible contraction factor $\rho'$ resulting form this parameterization, and compared it to the ones we achieve with the Lyapunov function defined in \Cref{thm:ef21}. In order to account for the possibility that there may be a better weighting between the two terms of \cref{eq:lyapunov_richtarik}, we let that the weighting be a free parameter in the Lyapunov analysis PEP.

\begin{figure}[h]
    \centering
    \includegraphics[width=\textwidth]{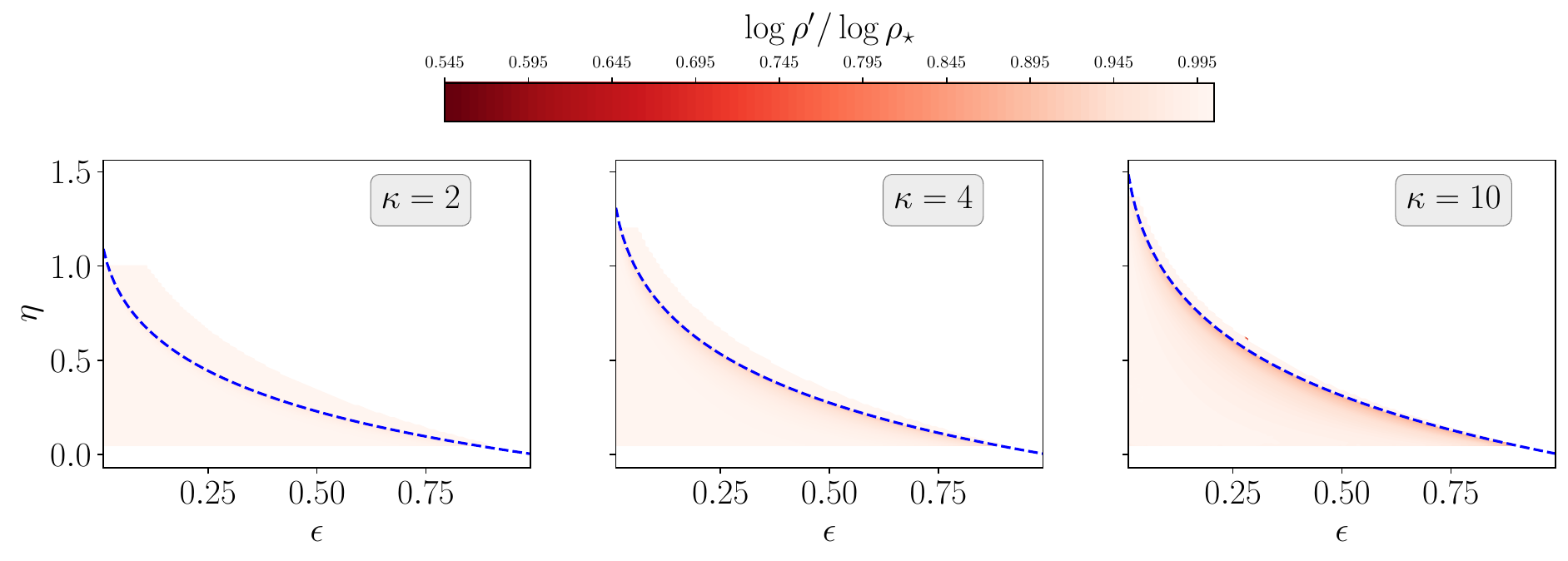}
    \caption{Complexity ratio between the rate achieved by the freely weighted version of \cref{eq:lyapunov_richtarik}, and the Lyapunov function used in \Cref{thm:ef21}.}
    \label{fig:lyapunov_comparison}
\end{figure}

The metric used is the complexity metric introduced in the beginning of \Cref{sec:rate_comparison}. The result of this comparison can be found in \Cref{fig:lyapunov_comparison}.

\clearpage

\section{Remarks on the importance of proof certificates} \label{sec:proof_certs}
The paper and appendix provide machine-checkable \emph{certificates} for the main 
claims in the paper. We include two such complementary certificate types:
\begin{enumerate}
    \item[{\includegraphics[height=1em,align=c]{CAS.pdf}}] \textbf{Symbolic certificates (WolframScript). } Short WolframScript snippets that verify algebraic identities and intermediary equalities used within proofs. Any suitable Computer Algebra System (CAS) could be used for these purposes.
    \item[{\includegraphics[height=1em, align=c]{PEP.pdf}}] \textbf{Numerical certificates (PEPs). } Instances of the PEPs (as defined in the main text) that validate global inequalities and performance bounds by directly solving the corresponding optimization problems. 
\end{enumerate}
While the proofs resulting from PEP formulations are well suited to CAS for local verification of individual steps, we use both tools: WolframScript checks the algebraic steps, and PEPs validate the global statements.

The rationale behind the decision of adding these proof certificates is the following:
\begin{itemize}
\item \textbf{Reviewing efficiency and literature consolidation.} Submission volumes at large ML venues continue to grow, increasing reviewer workload\footnote{\url{https://blog.neurips.cc/category/2024-conference/}} and raising challenges in the community, as acceptance of a paper does not necessarily guarantee that proofs have been thoroughly checked \footnote{The NeurIPS 2025 reviewer guidelines indicate specifically that the reviewer is \textit{not} required to read the paper's supplementary material \url{https://neurips.cc/Conferences/2025/ReviewerGuidelines}. }.

While machine-checkable certificates do not replace mathematical proofs, they may shift effort from re-deriving statements to verifying that appropriate checks were provided, which is typically easier. They also provide authors with early feedback on correctness before submission.
\item \textbf{Incremental path to formalization.} General-purpose proof assistants (e.g., Lean, Rocq) do not yet provide all mathematical infrastructure needed to formalize many ML results. As an interim step, we certify the parts that \emph{are} currently amenable: symbolic steps via computer algebra and end-to-end statements via PEPs. This may enable a gradual transition toward fully formal proofs as libraries mature.
\item \textbf{Clarifying experimental intent.} Ultimately, we also believe that this can serve to clarify the role of experiments in the papers, that are not always described. Some experiments test hypotheses; while others function as \emph{theory unit tests}. Declaring the latter explicitly (and supplying their certificates) helps readers understand how these checks support the claims.
\end{itemize}

Formalization in Lean or Rocq is our ultimate standard for a guaranteed, machine-checked certificate of validity. We consider this an ambition for future work and aim to explore it by gradually linking the symbolic components to Lean (e.g., turning recurring identities into lemmas) and packaging numerical outputs as checkable witnesses. 



\ifincludesection
\clearpage
\section*{NeurIPS Paper Checklist}

The checklist is designed to encourage best practices for responsible machine learning research, addressing issues of reproducibility, transparency, research ethics, and societal impact. Do not remove the checklist: {\bf The papers not including the checklist will be desk rejected.} The checklist should follow the references and follow the (optional) supplemental material.  The checklist does NOT count towards the page
limit. 

Please read the checklist guidelines carefully for information on how to answer these questions. For each question in the checklist:
\begin{itemize}
    \item You should answer \answerYes{}, \answerNo{}, or \answerNA{}.
    \item \answerNA{} means either that the question is Not Applicable for that particular paper or the relevant information is Not Available.
    \item Please provide a short (1–2 sentence) justification right after your answer (even for NA). 
\end{itemize}

{\bf The checklist answers are an integral part of your paper submission.} They are visible to the reviewers, area chairs, senior area chairs, and ethics reviewers. You will be asked to also include it (after eventual revisions) with the final version of your paper, and its final version will be published with the paper.

The reviewers of your paper will be asked to use the checklist as one of the factors in their evaluation. While "\answerYes{}" is generally preferable to "\answerNo{}", it is perfectly acceptable to answer "\answerNo{}" provided a proper justification is given (e.g., "error bars are not reported because it would be too computationally expensive" or "we were unable to find the license for the dataset we used"). In general, answering "\answerNo{}" or "\answerNA{}" is not grounds for rejection. While the questions are phrased in a binary way, we acknowledge that the true answer is often more nuanced, so please just use your best judgment and write a justification to elaborate. All supporting evidence can appear either in the main paper or the supplemental material, provided in appendix. If you answer \answerYes{} to a question, in the justification please point to the section(s) where related material for the question can be found.

IMPORTANT, please:
\begin{itemize}
    \item {\bf Delete this instruction block, but keep the section heading ``NeurIPS Paper Checklist"},
    \item  {\bf Keep the checklist subsection headings, questions/answers and guidelines below.}
    \item {\bf Do not modify the questions and only use the provided macros for your answers}.
\end{itemize}


\begin{enumerate}

\item {\bf Claims}
    \item[] Question: Do the main claims made in the abstract and introduction accurately reflect the paper's contributions and scope?
    \item[] Answer: \answerYes{} 
    \item[] Justification: Every claim is proven in the paper, and we make the setting
    and assumptions clear in the paper.

\item {\bf Limitations}
    \item[] Question: Does the paper discuss the limitations of the work performed by the authors?
    \item[] Answer: \answerYes{} 
    \item[] Justification: We discuss the fact that we are restricted to the single-agent
    setting in the discussion. We also make it quite clear that we are working
    with deterministic compression operators and smooth strongly convex functions in the background section.

\item {\bf Theory assumptions and proofs}
    \item[] Question: For each theoretical result, does the paper provide the full set of assumptions and a complete (and correct) proof?
    \item[] Answer: \answerYes{} 
    \item[] Justification: All assumptions and full theorem statements are in the paper, and the proofs
    are in the appendix.

    \item {\bf Experimental result reproducibility}
    \item[] Question: Does the paper fully disclose all the information needed to reproduce the main experimental results of the paper to the extent that it affects the main claims and/or conclusions of the paper (regardless of whether the code and data are provided or not)?
    \item[] Answer: \answerYes{} 
    \item[] Justification: Full details about how the plots were generated are given in the appendix. We also
    provide the code as part of the submission for the reviewers, and intend on making it open source afterwards.

\item {\bf Open access to data and code}
    \item[] Question: Does the paper provide open access to the data and code, with sufficient instructions to faithfully reproduce the main experimental results, as described in supplemental material?
    \item[] Answer: \answerYes{} 
    \item[] Justification: The code will be made open source after the review process. In the meantime,
    the reviewers will get access to the code in an anonymized form.
    \item[] Guidelines:
    \begin{itemize}
        \item The answer NA means that paper does not include experiments requiring code.
        \item Please see the NeurIPS code and data submission guidelines (\url{https://nips.cc/public/guides/CodeSubmissionPolicy}) for more details.
        \item While we encourage the release of code and data, we understand that this might not be possible, so "No" is an acceptable answer. Papers cannot be rejected simply for not including code, unless this is central to the contribution (e.g., for a new open-source benchmark).
        \item The instructions should contain the exact command and environment needed to run to reproduce the results. See the NeurIPS code and data submission guidelines (\url{https://nips.cc/public/guides/CodeSubmissionPolicy}) for more details.
        \item The authors should provide instructions on data access and preparation, including how to access the raw data, preprocessed data, intermediate data, and generated data, etc.
        \item The authors should provide scripts to reproduce all experimental results for the new proposed method and baselines. If only a subset of experiments are reproducible, they should state which ones are omitted from the script and why.
        \item At submission time, to preserve anonymity, the authors should release anonymized versions (if applicable).
        \item Providing as much information as possible in supplemental material (appended to the paper) is recommended, but including URLs to data and code is permitted.
    \end{itemize}

\item {\bf Experimental setting/details}
    \item[] Question: Does the paper specify all the training and test details (e.g., data splits, hyperparameters, how they were chosen, type of optimizer, etc.) necessary to understand the results?
    \item[] Answer: \answerYes{} 
    \item[] Justification: Necessary details to understand the plots are given in the captions.
    Details for how they were generated are given in the appendix.

\item {\bf Experiment statistical significance}
    \item[] Question: Does the paper report error bars suitably and correctly defined or other appropriate information about the statistical significance of the experiments?
    \item[] Answer: \answerYes{} 
    \item[] Justification: The paper does include experiments, but they are not random and
    so this question is not applicable.

\item {\bf Experiments compute resources}
    \item[] Question: For each experiment, does the paper provide sufficient information on the computer resources (type of compute workers, memory, time of execution) needed to reproduce the experiments?
    \item[] Answer: \answerNo{} 
    \item[] Justification: The experiments can be scaled arbitrarily to a fine or course grid
    of points, and the time to finish is highly dependent on this. Nevertheless, all the experiments
    were run on an M4 Max.
    
\item {\bf Code of ethics}
    \item[] Question: Does the research conducted in the paper conform, in every respect, with the NeurIPS Code of Ethics \url{https://neurips.cc/public/EthicsGuidelines}?
    \item[] Answer: \answerYes{} 
    \item[] Justification: The authors have not violated the code of ethics.

\item {\bf Broader impacts}
    \item[] Question: Does the paper discuss both potential positive societal impacts and negative societal impacts of the work performed?
    \item[] Answer: \answerNo{} 
    \item[] Justification: The paper only provides tight analysis of already existing methods.
    The only impact it will have is on the community of researchers working in distributed
    optimization.
    
\item {\bf Safeguards}
    \item[] Question: Does the paper describe safeguards that have been put in place for responsible release of data or models that have a high risk for misuse (e.g., pretrained language models, image generators, or scraped datasets)?
    \item[] Answer: \answerNA{} 
    \item[] Justification: No data or models are released in this paper.

\item {\bf Licenses for existing assets}
    \item[] Question: Are the creators or original owners of assets (e.g., code, data, models), used in the paper, properly credited and are the license and terms of use explicitly mentioned and properly respected?
    \item[] Answer: \answerYes{} 
    \item[] Justification: Citations for the software packages PEPit and PySR are present in paper.

\item {\bf New assets}
    \item[] Question: Are new assets introduced in the paper well documented and is the documentation provided alongside the assets?
    \item[] Answer: \answerYes{} 
    \item[] Justification: The only new asset introduced is the code implementing the 
    feasibility problems themselves, and the utility functions surrounding them.
    The documentation for these are given in the form of instructions for reviewers
    to replicate our findings in the supplementary material, and in the comments
    of the code.

\item {\bf Crowdsourcing and research with human subjects}
    \item[] Question: For crowdsourcing experiments and research with human subjects, does the paper include the full text of instructions given to participants and screenshots, if applicable, as well as details about compensation (if any)? 
    \item[] Answer: \answerNA{} 
    \item[] Justification: No crowdsourcing or human subjects are involved in this paper.

\item {\bf Institutional review board (IRB) approvals or equivalent for research with human subjects}
    \item[] Question: Does the paper describe potential risks incurred by study participants, whether such risks were disclosed to the subjects, and whether Institutional Review Board (IRB) approvals (or an equivalent approval/review based on the requirements of your country or institution) were obtained?
    \item[] Answer: \answerNA{} 
    \item[] Justification: No research with human subjects is involved in this paper.

\item {\bf Declaration of LLM usage}
    \item[] Question: Does the paper describe the usage of LLMs if it is an important, original, or non-standard component of the core methods in this research? Note that if the LLM is used only for writing, editing, or formatting purposes and does not impact the core methodology, scientific rigorousness, or originality of the research, declaration is not required.
    \item[] Answer: \answerNA{} 
    \item[] Justification: The paper does not use LLMs as any important, original, or non-standard components.

\end{enumerate}
\fi
\end{document}